%% file: main.tex
\documentclass[10pt]{article} 
\usepackage[preprint]{rlc}

\usepackage{amssymb}            
\usepackage{mathtools}          
\usepackage{mathrsfs}           
\usepackage{graphicx}           
\usepackage{subcaption}         
\usepackage[space]{grffile}     
\usepackage{url}                

\title{Policy Gradient Algorithms with Monte Carlo Tree Learning for Non-Markov Decision Processes}

\author{Tetsuro Morimura  \\
    morimura\_tetsuro@cyberagent.co.jp \\
    CyberAgent\\
    \And
    Kazuhiro Ota \\
    CyberAgent\\
    \And
    Kenshi Abe \\
    CyberAgent \\
    \And
    Peinan Zhang \\
    CyberAgent
}

\usepackage{algorithm}
\usepackage{algpseudocode}

\usepackage{bm}
\usepackage{sidecap}

\usepackage{tablefootnote}  
\usepackage{threeparttable}

\usepackage{setspace}

\usepackage{amsthm}
\usepackage{booktabs}

\newtheorem{theorem}{Theorem}
\newtheorem{proposition}{Proposition}
\newtheorem{lemma}{Lemma}

\newtheorem{assumption}{Assumption}

\newif\ifpreprint
\preprintfalse    

\input{def.tex}

\renewcommand{\cite}{\citep}

\begin{document}

\maketitle

\begin{abstract}
Policy gradient (PG) is a reinforcement learning (RL) approach that optimizes a parameterized policy model for an expected return using gradient ascent. While PG can work well even in non-Markovian environments, it may encounter plateaus or peakiness issues. As another successful RL approach, algorithms based on Monte Carlo Tree Search (MCTS), which include AlphaZero, have obtained groundbreaking results, especially in the game-playing domain. They are also effective when applied to non-Markov decision processes. However, the standard MCTS is a method for decision-time planning, which differs from the online RL setting. In this work, we first introduce Monte Carlo Tree Learning (MCTL), an adaptation of MCTS for online RL setups. We then explore a combined policy approach of PG and MCTL to leverage their strengths. We derive conditions for asymptotic convergence with the results of a two-timescale stochastic approximation and propose an algorithm that satisfies these conditions and converges to a reasonable solution. Our numerical experiments validate the effectiveness of the proposed methods.
\end{abstract}

\input{sec_intro.tex}
\input{sec_background.tex}
\input{sec_proposed_method.tex}
\input{sec_related_work.tex}
\input{sec_experiments.tex}

\input{sec_conclusion.tex}

\bibliography{refs}
\bibliographystyle{rlc}

\appendix

\input{supplement.tex}

\end{document}

%% file: def.tex
\newcommand{\calA} {\mathcal{A}}

\newcommand{\calF} {\mathcal{F}}
\newcommand{\calH} {\mathcal{H}}
\newcommand{\calO} {\mathcal{O}}

\newcommand{\calS} {\mathcal{S}}

\newcommand{\bbN}  {\mathbb{N}}
\newcommand{\bbR}  {\mathbb{R}}

\newcommand{\Proof}{\noindent{{\bf Proof:\ \ }}}

\newcommand{\romani}  {$\rm(\hspace{.18em}i\hspace{.18em})$}
\newcommand{\romanii} {$\rm(\hspace{.08em}ii\hspace{.08em})$}
\newcommand{\romaniii}{$\rm(i\hspace{-.08em}i\hspace{-.08em}i)$}
\newcommand{\romaniv} {$\rm(i\hspace{-.08em}v\hspace{-.06em})$}

\newcommand{\given}    {\,\vert\,}

\newcommand{\abs}[1]   {\lvert#1\rvert}
\newcommand{\norm}[1]  {\lVert#1\rVert}
\newcommand{\normlr}[1] {\left\lVert#1\right\rVert}

\newcommand{\transs}   {\top}			
\newcommand{\trans}    {^{\!\transs}}		

\newcommand{\Env}     {\mathsf{HDP}}
\newcommand{\pini}    {p_{\textrm{ini}}}

\newcommand{\rfun}    {f_{\textrm{r}}}
\newcommand{\po}      {p_{\textrm{o}}}
\newcommand{\ph}      {p_{\textrm{h}}}
\newcommand{\ret}     {g} 
\newcommand{\Ret}     {G} 
\newcommand{\obj}     {\Upsilon} 

\newcommand{\rmax}    {{R_{\rm max}}}
\ifpreprint 
\newcommand{\tmax}    {{T_{\rm max}}}
\else
\newcommand{\tmax}    {{T}}

\newcommand{\polx}    {\pi^{\hspace{-0.2mm}\xchar\hspace{-0mm}}}
\newcommand{\polxn}[1][n]   {\pi^{\hspace{-0.2mm}\xchar_{\hspace{-0.3mm}#1}\hspace{-0.4mm}}}
\newcommand{\polxt}[1][\tau]{\pi^{\hspace{-0.2mm}\xchar{\hspace{-0.1mm}(\hspace{-0.2mm}#1\hspace{-0.1mm})}\hspace{-0.3mm}}}

\newcommand{\poly}    {\pi^{\hspace{-0.2mm}\ychar\hspace{-0mm}}}
\newcommand{\polyn}[1][n]   {\pi^{\hspace{-0.1mm}\ychar_{\hspace{-0.3mm}#1}\hspace{-0.4mm}}}
\newcommand{\polxy}   {\pi^{\xchar\!,\ychar}}

\newcommand{\polxyn}[1][n] {\pi^{\xchar_{\hspace{-0.2mm}#1}\hspace{-0.4mm},\ychar_{\hspace{-0.1mm}#1}}}

\newcommand{\E}        {\mathbb{E}}
\newcommand{\Epol}     {\mathbb{E}^{\pi\hspace{-0.2mm}}}
\newcommand{\Epolx}    {\mathbb{E}^{\polx\hspace{-0.2mm}}}
\newcommand{\Epolxn}   {\mathbb{E}^{\polxn\hspace{-0.2mm}}}

\newcommand{\Epolxy}   {\mathbb{E}^{\polxy\hspace{-0.2mm}}}
\newcommand{\Epolxyn}  {\mathbb{E}^{\polxyn\hspace{-0.2mm}}}

\newcommand{\visitprob}  {d^{\hspace{0.2mm}\xchar\!,\ychar\hspace{-0.3mm}}}

\newcommand{\qhat}   {q}        
\newcommand{\invvst} {u}        

\newcommand{\xchar}   {\theta}
\newcommand{\ychar}   {\omega}
\newcommand{\zchar}   {\theta}
\newcommand{\x}       {\xchar}
\newcommand{\y}       {\ychar}
\newcommand{\z}       {\zchar}

\newcommand{\xn}[1][n]{\x_{#1}}
\newcommand{\yn}[1][n]{\y_{#1}}
\newcommand{\zn}[1][n]{\z_{#1}}
\newcommand{\zs}[1][s]{\z^{#1\hspace{-0.3mm}}}

\newcommand{\xlr}     {\alpha}
\newcommand{\ylr}     {\eta}
\newcommand{\zlr}     {\alpha}
\newcommand{\xmap}    {k}
\newcommand{\xmapp}   {\tilde{k}}
\newcommand{\ymap}    {l}
\newcommand{\zmap}    {k}
\newcommand{\xmds}    {M^{(1)}}
\newcommand{\ymds}    {M^{(2)}}
\newcommand{\zmds}    {M}
\newcommand{\mds}[1]  {M^{(#1)}}
\newcommand{\xbias}   {\epsilon^{(1)}}
\newcommand{\ybias}   {\epsilon^{(2)}}
\newcommand{\zbias}   {\epsilon}
\newcommand{\bias}[1] {\epsilon^{(#1)}}

\newcommand{\difx}    {\nabla_{\hspace{-0.7mm}\xchar\hspace{0.1mm}}}

\newcommand{\mfun}       {\lambda}
\newcommand{\mfunn}[1][n]{\lambda_{#1}}

\newcommand{\T}{\nu}

%% file: sec_intro.tex
\section{Introduction}
\label{sec:intro}
Reinforcement learning (RL) attempts to learn a policy model 
so as to maximize the average of cumulative rewards \cite{sutton2nd}.
Policy gradient (PG) algorithms employ gradient ascent on policy parameters
\cite{gullapalli90a,williams92a,baxter01a}.
They can benefit much from recent advances in neural network models
and have been applied in various challenging domains,
\ifpreprint
such as robotics \cite{Peters08b}, image captioning \cite{SCST}, document summarization \cite{Paulus18a}, 
large language models \cite{instructGPT}
and speech recognition \cite{Zhou18a}.
\else
such as robotics \cite{Peters08b}, text generation \cite{SCST,instructGPT},
and speech recognition \cite{Zhou18a}.
\fi

Monte Carlo Tree Search (MCTS) is another successful RL approach, 
combining Monte Carlo sampling with an optimistic tree search 
that balances exploration and exploitation \cite{Kocsis06a,Coulom06a,Browne12a}. 
Notably, when integrated with deep learning, as in AlphaZero \cite{Silver17a,Silver17b}
and MuZero \cite{Schrittwieser20a}, 
MCTS algorithms have achieved groundbreaking results in board games \cite{Silver16a}.

%

Ordinary RL assumes the environment has the Markov property,
i.e., the reward process and system dynamics of the underlying process are Markovian.
More specifically, they depend only on the current state (and action);
in other words, given the current state, they are independent of the past states.
It enables computationally effective dynamic programming techniques
to learn policy models \cite{Puterman94a,bertsekas95a}.
However, in many real-world RL tasks,
it is difficult to determine in advance a good state set or space that satisfies the Markov property
\cite{Yu11a,Friedrich11a,Berg12a,Clarke15a,SCST,Paulus18a,Zhou18a,You18a}.

There are at least two typical scenarios where the Markov property is violated.
The first is related to the observation.
If observations are limited and partial,
the dynamics and rewards are not Markovian
and need to be modeled with functions of the past observation sequence or functions of a latent state.
Typical examples are dialog systems \cite{Young13a} and robot navigation \cite{Berg12a}.
The other case is when only the reward function is not Markovian.
Generation tasks, such as text \cite{SeqGAN} and molecular graphs \cite{You18a},
are a typical examples
since generated objects are usually evaluated not only from a local but also from a global perspective,
such as an ad-quality score in the domain of text generation for search engine advertising \cite{Kamigaito21a}.
The former scenario is often formulated as a partially observable Markov decision process (POMDP)
\cite{kaelbling96a,sondikthesis}, 
while the latter is a decision process with non-Markovian reward 
 \cite{Bacchus96a}.
The stochastic process that includes both is called a non-Markovian decision process (NMDP) 
or history-based decision process (HDP) \cite{Whitehead95a,Bacchus97a,Majeed18a},
which is the focus of this paper.

\begin{figure}[t]
 \centering
 \includegraphics[width=4.5in]{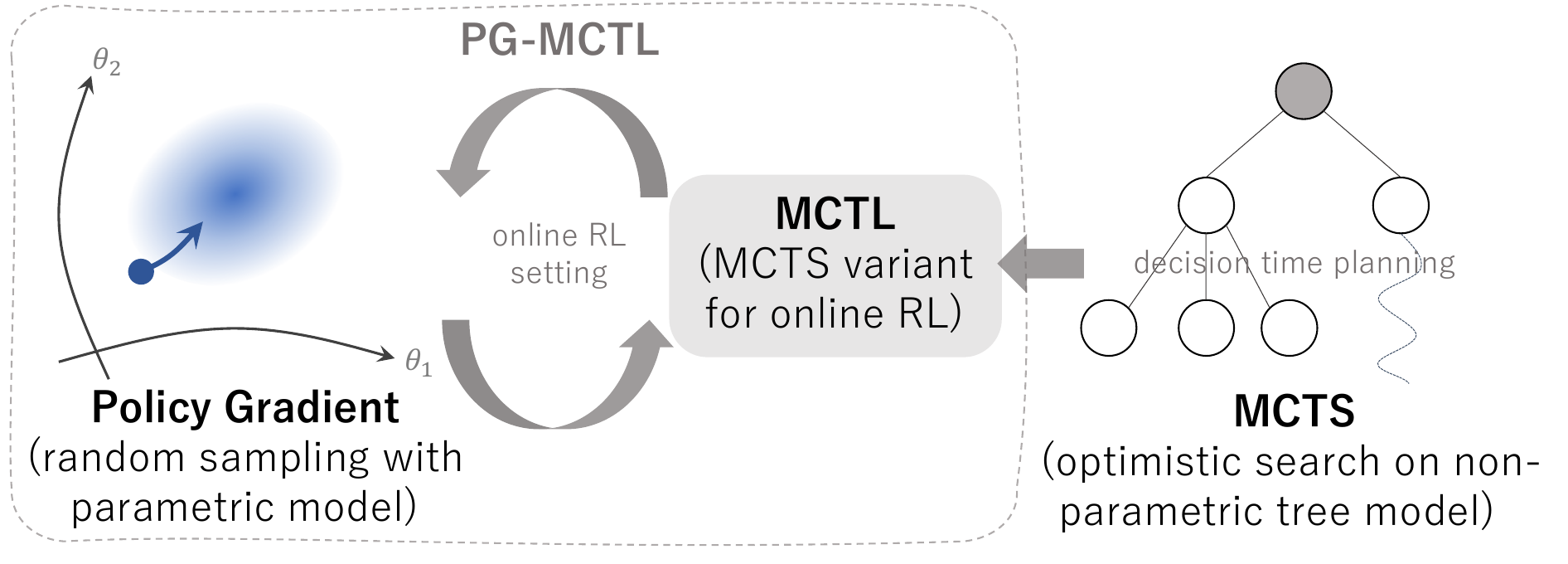}
 \vspace{-1mm}
 \caption{
 Overview of the proposed approach;\hspace{-0.2mm} \textit{P\hspace{-0.1mm}G guided by Monte\hspace{-0.1mm} Carlo\hspace{-0.1mm} Tree\hspace{-0.1mm} Learning\hspace{-0.1mm} (PG-MCTL).}
 Unlike MCTS, which requires a simulator to generate possible future states and rewards, MCTL builds a tree based on real trajectories experienced by an agent while still inheriting core MCTS properties.
 PG and MCTL have fundamentally different properties. PG-MCTL takes advantage of them.
 }
 \label{fig:concept}
 \vspace{-2mm}
\end{figure}

Notably, both PG and MCTS algorithms are applicable to HDP-modeled tasks 
\cite{kimura97a, AberdeenThesis, SCST, Browne12a},
as they are less reliant on the Bellman optimality equation under the Markov assumption, unlike Q learning.
Moreover, PG algorithms can effectively utilize function approximators like neural networks.
%
%
%
However, PGs are known to occasionally get trapped on plateaus, slowing down learning
\cite{kakade02a,morimura14:aaai,Ciosek20a}.
%
Furthermore, PGs can face the 'peakiness' issue, where
the initially most probable actions will gain probability mass, 
even if they are not the most rewarding \cite{Choshen20a,Kiegeland21a}.
Meanwhile, MCTS-based algorithms aim for a global optimum through optimistic search,
balancing exploration and exploitation \cite{Kocsis06a,Lattimore20a,Swiechowski21a}. 
Yet, compared to PGs, they struggle with state generalization,
often lacking information on states outside of their tree.
%
%
%
Furthermore, while PGs do not require a simulator for action selection, 
MCTSs do.
That is, they are decision-time planning \cite{sutton2nd}, 
in which planning is launched and completed for every action selection.

Based on the above, we believe that PG and MCTS can complement each other's difficulties,
and their combination is a promising way to solve problems in HDPs.
Specifically, 
even when PG is suffering from plateaus or the peakiness issue, 
MCTS is likely to be able to continue improving
because its exploration strategy is fundamentally different from PG's.
In addition, PG with an appropriately parameterized model would be able
to cover the inefficiency in the state generalization of MCTS.
It is also generally known that a combination of models can have a positive effect \cite{Kuncheva14a}.

This paper considers an online model-free RL problem in HDPs, where the environment will not be estimated. That is, no simulator is available.
We adapt MCTS to the online RL setup 
and propose Monte Carlo Tree Learning (MCTL) that selects an action without a simulator.
We then consider an approach that uses a mixture of PG and MCTL policies and adjusts its mixing probability through learning.
%
We call this approach {\it a policy gradient guided by MCTL} (PG-MCTL)
(Figure \ref{fig:concept}).
We derive conditions for asymptotic convergence 
and find that naive mixing of PG and MCTL will not work in asymptotic convergence.
We propose an algorithm that satisfies the conditions for convergence.

This paper is structured as follows: 
Section \ref{sec:background} covers the background of RL in HDPs, PG, and MCTS.
In Section \ref{sec:proposed_method}, we introduce the PG-MCTL approach,
detail its convergence conditions using a two-timescale stochastic approximation,
and present an implementation that meets these conditions. This is our main contribution. 
Section \ref{sec:related_work} reviews relevant literature.
The effectiveness of the proposed approach is validated through experiments in Section \ref{sec:experiments}, 
and Section \ref{sec:conclusion} offers concluding remarks.



%% file: sec_background.tex
\section{Preliminaries}
\label{sec:background}

We define our problem setting of RL in HDPs in Section \ref{subsec:problem_setting}.
PG and MCTS algorithms are briefly reviewed in Sections \ref{subsec:pg} and \ref{subsec:mcts}, respectively.

\subsection{Problem setting of RL in HDP}
\label{subsec:problem_setting}

While problems of RL are usually formulated on a Markov decision process (MDP) for ease of learning \cite{sutton2nd},
as described in Section \ref{sec:intro}, 
it is difficult to define Markovian states in many real-world tasks.
Here, we consider 
a discrete-time episodic HDP \cite{Whitehead95a, Majeed18a}
as a general decision process without assuming the Markovian property.
It is defined by a tuple
$\Env \triangleq \{\calO, \calA, \tmax, \pini, \po, \rfun\}$,
\ifpreprint
which are as follows:
\begin{itemize}
 \item $\calO$ is a finite set of observations.
 \item $\calA$ is a finite set of actions.
 \item $\tmax$ is the length of each episode.
 \item $\pini:\calO\rightarrow[0,1]$ is a probability function of the initial observation,
       $\pini(o_0) \triangleq \Pr(o_0)$
       \footnote{
       Although it should be $\Pr(O_0\!=\!o_0)$ for the random variable $O_0$ and realization $o_0$ to be precise,
       we write $\Pr(o_0)$ for brevity.
       The same rule is applied to the other probability functions if there is no confusion.
       }.
 \item $\po:\calO\times\calH_t\times\calA \rightarrow [0,1]$ is
       a history-dependent observation probability function
       at each time step $t\in\{0,1,\dots,\tmax\}$,
       \begin{align}
	\hspace{-1mm}
	\po(o_{t+1}| h_t, a_t) &\triangleq \Pr(O_{t+1}\!\!=\!o_{t+1} \given H_t\!=\!h_t,A_t\!=\!a_t),
	\ \ \ \forall (o_{t+1},h_t,a_t) \in \calO\times\calH_t\times\calA,
	\notag
       \end{align}              
       where \mbox{$h_t\!\triangleq\![o_0,a_0,\dots,o_{t-1},a_{t-1},o_t]\!=\![h_{t-1},a_{t-1},o_t]$}
       is a history up to a time step $t$, and
       \mbox{$\calH_t \triangleq (\calO\times\calA)^t \times\calO$}  is a set of histories at a time step $t$.
       We notate the total history set $\calH\triangleq\bigcup_{t=0}^{\tmax}\calH_t$
       and the history transition probability function
       $\ph:\calH_{t+1}\times\calH_t\times\calA \rightarrow [0,1]$ 
       at $t\in\{0,\dots,\tmax\!-1\}$ such as
       \begin{align}
	\ph(h_{t+1}\!=\![h_t,a_t,o_{t+1}] \given h_t,a_t) \triangleq \po(o_{t+1}|h_t,a_t).
	\notag
       \end{align}
 \item $\rfun:\calH\times\calA \rightarrow [-\rmax,\rmax]$ is a history-dependent bounded
       reward function, which defines an immediate reward $r_t = \rfun(h_t,a_t)$ 
       at time step $t\in\{0,\dots,\tmax\}$.
\end{itemize}
Here we bring in the length $\tmax$ mainly to simplify the presentation of the paper.
If we set $\tmax$ to a sufficiently large value and assume that once an agent reaches a terminal state,
it stays there until $\tmax$,
the effect of $\tmax$ can be practically ignored.
\else 
where
$\calO$ and $\calA$ are finite sets of observations and actions, respectively.
$\tmax$ is the length of each episode,
$\pini:\calO\rightarrow[0,1]$ is a probability function of the initial observation,
       \mbox{$\pini(o_0)\triangleq\Pr(o_0)$}%
       \footnote{
       Although it should be $\Pr(O_0\!=\!o_0)$ for the random variable $O_0$ and realization $o_0$ to be precise,
       we write $\Pr(o_0)$ for brevity.
       The same rule is applied to the other probability functions if there is no confusion.
       }, and
$\po:\calO\times\calH_t\times\calA \rightarrow [0,1]$ is
       a history-dependent observation probability function
       at each time step $t\!\in\!\{0,1,\dots,\tmax\!-\!1\}$,
       \mbox{$\po(o_{t+1} \!\given h_t, a_t)\!\triangleq\! \Pr(O_{t+1}\!\!=\!o_{t+1} \!\given H_t\!=\!h_t,A_t\!=\!a_t)$},
       where \mbox{$h_t\!\triangleq\![o_0,a_0,\dots,o_{t-1},a_{t-1},o_t]\!=\![h_{t-1},a_{t-1},o_t]$}
       is a history up to a time step $t$, and
       \mbox{$\calH_t \triangleq (\calO\times\calA)^t \times\calO$}  is a set of histories at a time step $t$.
       For brevity, we notate the total history set $\calH\triangleq\bigcup_{t=0}^{\tmax}\calH_t$
       and the history transition probability function
       $\ph:\calH_{t+1}\times\calH_t\times\calA \rightarrow [0,1]$ 
       such as
       $
	\ph(h_{t+1}\!=\![h_t,a_t,o_{t+1}] \given h_t,a_t) \triangleq \po(o_{t+1}|h_t,a_t)
	$.
The function $\rfun\!:\!\calH\times\calA \rightarrow \bbR$ is a history-dependent bounded
       reward function, which defines an immediate reward $r_t = \rfun(h_t,a_t)$ 
       at time step $t\in\{0,\dots,\tmax\}$.
\fi

A learning agent chooses an action according to a policy model
\mbox{$\pi:\calA\times\calH\rightarrow[0,1]$}, which is a conditional action probability function 
$\pi(a|h_t)\!\triangleq\!\Pr(a\given h_t, \pi)$ 
at each time step $t$.
%
%
Without loss of generality, we assume that the agent can take 
any action $a\in\calA$ in any $h_t\in\calH_t$ at any $t \in \{0,\dots,\tmax\}$.

\ifpreprint 
Here, we consider a standard online RL problem, where 
$\pini$, $\po$, and $\rfun$ are unknown to the agent
\footnote{Even when they are known to the agent, however it is almost impossible to analytically evaluate Eq.~\!\eqref{eq:obj} or solve RL problems in HDPs. This is because there is a combinatorially large number of histories
and thus the history space cannot be fully searched. So an approximation approach like RL is necessary in HDPs.}. 
The agent learns the policy model $\pi$ by experiencing episodes repeatedly.
The objective function that the agent seeks to maximize is the expected return 
\begin{align}
 \obj(\pi) \triangleq \Epol[ G_0 ],
 \label{eq:obj}
\end{align}
where we use this notation $\Epol[\,\cdot\,]\triangleq \E[\,\cdot\given \Env, \pi]$ and 
$G_t$ is a random variable of the return at time step $t \in \{0,\dots,\tmax\}$,
\begin{align}
 G_t \triangleq \sum_{k=t}^\tmax R_k = \sum_{k=t}^\tmax \rfun(H_k,A_k).
 \notag 
\end{align}
Although the discount factor $\gamma\in[0,1]$ is often used,
as in $G_t = \sum_{k=t}^{\tmax} \gamma^{k-t} R_k$, 
we omit it for simplicity, since all our results are immediately applicable to the case with $\gamma<1$.

\else 
Here, we consider a standard online RL problem, where 
$\pini$, $\po$, and $\rfun$ are unknown to the agent. 
The agent learns the policy model $\pi$ by experiencing episodes repeatedly.
The objective function that the agent seeks to maximize is the expected return 
\begin{align}
 \obj(\pi) \triangleq \Epol[ G_0 ],
 \label{eq:obj}
\end{align}
where 
$\Epol[\,\cdot\,]\triangleq \E[\,\cdot\given \Env, \pi]$ is the expectation operator and 
$G_t\!\triangleq\!\sum_{\kappa=t}^\tmax R_\kappa \!=\!\sum_{\kappa=t}^\tmax \rfun(H_\kappa,A_\kappa)$ is a random variable of the return at time step $t$. 
%
\fi 

\subsection{Policy gradient}
\label{subsec:pg}

We assume that the policy model $\polx$ to be optimized by PG algorithms
is parameterized by a parameter $\x\in\bbR^{d}$ 
and $\polx$ is differentiable with respect to $\x$. 
\ifpreprint
A typical instance of $\polx$ 
is a neural network for sequence learning, 
such as a recurrent neural network (RNN) \cite{Hochreiter97a,Wierstra10a,SCST}
and Transformer \cite{Vaswani17a,Chen21a}.
\else
Examples of $\polx$ include a neural network for sequence modeling.
\fi
The PGs are based on the gradient method of the following update rule
with a small learning rate $\xlr\geq0$,
\begin{align}
 \x &:= \x + \xlr \difx \obj(\polx), 
 \notag
\end{align}
where $:=$ is the right-to-left substitution operator
and 
$\difx \obj(\polx)
 \triangleq [\partial \obj(\polx)/ \partial{\xchar_1},..., \partial \obj(\polx) /\partial{\xchar_d} ]\trans$
is the gradient of $\obj(\polx)$ with respect to $\x$.
%
Because the analytical evaluation of $\difx \obj(\polx)$ is generally intractable,
a PG method, called REINFORCE \cite{williams92a}, updates
 $\xn$ after every episode $n$ of experience $[o_0,a_0,r_0,\dots,o_\tmax,a_\tmax,r_\tmax]$
according to a stochastic gradient method as follows:
\begin{align}
 \xn[n+1] = \xn + \xlr_n \sum_{t=0}^\tmax \difx \log \polxn(a_t|h_t)\, (\ret_t - b(h_t)), 
 \label{eq:pg_update}
\end{align}
since the gradient $\difx \obj(\polx)$ is written as
\begin{align}
   \difx \obj(\polx) 
 = \Epolx \!\! \left[\, \sum_{t=0}^\tmax\difx \log \polx (A_t, H_t)\, (\Ret_t - b(H_t))  \right]\!\!,
 \notag
\end{align}
where $\ret_t$ is the realized value of the return $\Ret_t$
and \mbox{$b:\calH\rightarrow\bbR$} is an arbitrary baseline function.
\ifpreprint 
The baseline function $b$ is used for reducing the variance of the stochastic gradient,
which is the second term on the right side of Eq.~\!\eqref{eq:pg_update},
and does not induce any bias to the gradient as long as it does not depend on the action
because of
\begin{align*}
  \Epolx&[ \difx\log\polx(A_t| H_t)\, b(H_t) \given H_t\!=\!h ]
 = b(h) \sum_{a} \difx\polx(a|h)
 = b(h) \difx \sum_{a} \polx(a|h)
 = b(h) \difx 1 = 0.
\end{align*}
Here, we notate $\sum_{a\in\calA}$ as $\sum_a$ for simplicity.
\else  
The baseline function $b$ is used for reducing the variance of the stochastic gradient,
and does not induce any bias to the gradient
because of
$\Epolx\![ \difx\log\polx(A_t| h)\, b(h)]=  b(h) \difx \!\sum_{a\in\calA}\! \polx(a|h) = b(h) \difx 1 = 0$.
\fi

\subsection{Monte Carlo tree search}
\label{subsec:mcts}

Monte Carlo tree search (MCTS) is developed to identify the best action 
in a given situation for decision processes \cite{Kocsis06a,Coulom06a,Browne12a}.
It is typically employed for decision-time planning, 
where planning is initiated and completed per action selection. 

It is typically employed for decision-time planning,
where planning is initiated and completed per action selection with a simulator. 
Here, a simulator is an environment model that can generate possible future states and rewards given a current state and action.
This allows the algorithm to simulate different action sequences to plan the optimal policy.
In planning, MCTS iteratively runs an episode from a given situation 
and stores its result in a tree by expanding the tree and updating statistics in nodes of the tree.
After a certain number of iterations, it estimates the best action in the situation by using statistics of the root node and terminates the planning.

In single-agent learning in a stochastic system, the tree usually has two kinds of nodes, a history node and a history-action node,
alternating in the depth direction.
A history node represents a history and does not store additional information. 
At a history node, the tree-search policy determines which child history-action node to transition to, based on the statistics in its child nodes. 
%
In contrast, the transition from a history-action node to a history node follows the transition probability $\ph$. 
%

Each history-action node holds a return estimate $q$ and the number of visits $m$
as the statistics.
Here, we notate those statistics in each history-action node $(h,a)$ with a tabular representation,
as $q(h,a)$ and $m(h,a)$, for simplicity.
The tree policy at a history node $h$,
often utilizing the Upper Confidence Bounds applied for Trees (UCT) formula \cite{Kocsis06a}, 
decides an action based on the statistics of its child nodes:
\begin{align}
 \arg \max_{a} \left\{ q(h,a) + C \sqrt{\frac{\log(\sum_b m(h,b))}{m(h,a)}}\right\},
 \label{eq:uct}
\end{align}
where $C\geq0$ is a hyper-parameter to control the balance between exploration and exploitation.

Each iteration of the MCTS consists of four consecutive phases:
\vspace{-3.2mm}
\begin{itemize}
    \item[\romani] selection of child nodes from the root to a leaf node in the tree,
    \vspace{-1.6mm}
    \item[\romanii] tree expansion by creating new child nodes that are initialized as $m\!:=\!1,\ q\!:=\!0$,
    \vspace{-1.6mm}
    \item[\romaniii] simulation from one of the new nodes according to a default policy to sample a return,
    \vspace{-1.6mm}   
    \item[\romaniv] backpropagation of the results until the root node.
    \vspace{-3.2mm}
\end{itemize}
Note that \romanii\ and \romaniii\ are skipped if the leaf node reached is a terminal node, and \romaniii\  is the ``Monte Carlo'' part of the algorithm.
The default policy in \romaniii\ is usually a uniform random policy.
%

In the backpropagation phase of \romaniv, 
the statistics of the node visited at each depth $t$ of iteration $n\in\{1,2,\dots\}$ are updated as follows:
\begin{align}
 \begin{cases}
   m_{n+1}(h_t, a_t) 
   = m_n(h_t, a_t) + 1,
   \\
   q_{n+1}(h_t, a_t) 
   =  q_n(h_t, a_t) + \frac{1}{m_n(h_t,a_t)} (g_t - q_n(h_t,a_t)).
 \end{cases}
 \label{eq:mcts_update}
\end{align}
Note that many other update rules have been proposed, such as the TD($\lambda$) learning type
\cite{Browne12a,Vodopivec17a}.

%% file: sec_proposed_method.tex
\section{Policy gradient guided by MCTL}
\label{sec:proposed_method}
In Section \ref{subsec:mctl},
we introduce Monte Carlo Tree Learning (MCTL) as an MCTS variant for online RL.
We then outline our approach, PG guided by MCTL (PG-MCTL), in Section \ref{subsec:approach}.
\ifpreprint 
Convergence analysis of the PG-MCTL is provided in Section \ref{subsec:theoretical_analysis}.
In Section \ref{subsec:implementation}, we propose an implementation
that satisfies the convergence conditions
and converges to a reasonable solution.
\else 
Convergence analysis is discussed in Section \ref{subsec:theoretical_analysis}, 
followed by a convergent implementation proposal in Section \ref{subsec:implementation}.
\fi 

\subsection{Monte Carlo tree learning (MCTL)}
\label{subsec:mctl}

Our focus is an online model-free RL problem in HDPs, 
where no simulator is available.
An agent learns through interactions with an unknown environment without estimating it. 
The standard MCTS, however, typically necessitates a simulator.
In response, we propose a variant of MCTS that inherits its key features but eliminates the need for a simulator.
This is referred to as a lazy MCTS or simply, MCTL.

MCTL gradually grows a tree according to the trajectories experienced by the agent interacting with an unknown environment.
%
Unlike MCTS, MCTL maintains and updates a tree over multiple episodes.%
\footnote{In MCTS, the search tree is built and used within a single episode, often requiring a simulator to generate possible future states. In contrast, MCTL continuously updates a single tree over multiple episodes using only the states experienced by the agent. This allows MCTL to function without a simulator, as it relies solely on real-world interactions.}
Specifically, after experiencing an episode, the tree is updated according to 
MCTS's tree expansion and backpropagation procedures %
\ifpreprint (e.g.~Eq.~\eqref{eq:mcts_update}). \else (e.g.~Eq.~\eqref{eq:mcts_update}). \fi
Depending on the presence of a node for the current situation $h$ in the tree, 
the MCTL policy adjusts its action selection.
If an MCTL policy is queried on a history $h$ that the tree contains,
it selects an action according to the node selection procedure of MCTS (e.g.~Eq.~\eqref{eq:uct}).
Otherwise, it selects an action randomly.

By design, 
an MCTL tree will have nodes for states near the initial state and/or frequently visited states.
The presence of a node $h$ implies that the history $h$ is somewhat known to an MCTL policy.
In contrast, the absence of a node $h'$ implies that the history $h'$ is unknown and 
the policy should select an action for exploration at $h'$.
This is analogous to a common class of algorithms, {\it knows what it knows} (KWIK), for efficient exploration \cite{Li11a}.

We will notate an MCTL policy as $\poly$, whose parameter is $\y$.
Specific implementations, including update rules, are described in Section \ref{subsec:implementation}.

\subsection{General approach}
\label{subsec:approach}

As discussed in the introduction, 
combining PG and MCTL aims to leverage the strengths and mitigate the weaknesses of each individual method.
Our proposed approach, PG guided by MCTL (PG-MCTL), 
integrates them by randomly selecting a policy of either PG or MCTL at each time step. 
Specifically, we consider the following mixture of policies $\polx$ and $\poly$:
\begin{align}
 \polxy(a|h) \triangleq (1-\mfun(h)) \polx(a|h) + \mfun(h) \poly(a|h),
 \label{eq:polxy}
\end{align}
where $\mfun$ is a mixing probability.
The $\mfun$ may be constant or depend on an observation $o$, history $h$, and
parameters $\x$, $\y$.
The parameters $\x$ and $\y$ are updated with modified update rules of a PG and MCTL, respectively, 
which are proposed in Section \ref{subsec:implementation} by using the results in Section \ref{subsec:theoretical_analysis}.

\subsection{Convergence analysis}
\label{subsec:theoretical_analysis}

We present the convergence conditions of PG-MCTL 
on a few settings of the mixing probability $\mfunn$ in Eq.~\!\eqref{eq:polxy}.
%
Proofs are shown in Appendix \ref{appendix:proofs}.

%
We will validate our assumptions in the next section.
For now, let's assume 
the updates of parameters  $\x\in\bbR^{d}$ for a PG policy
and $\y\in\bbR^{e}$ for an MCTL policy can be rewritten as follows:
\begin{align}
 \xn[n+1] &= \xn + \xlr_n [\, \xmap(\xn,\yn) + \xmds_{n+1} + \xbias_n],
 \label{eq:x_update_n}
 \ifpreprint\\ \else \\[-0.5mm] \fi
 \yn[n+1] &= \yn + \ylr_n [\, \ymap(\xn,\yn) + \ymds_{n+1} + \ybias_n],
 \label{eq:y_update_n}
\end{align}
where $\xmap:\bbR^{d}\times\bbR^e\!\rightarrow\bbR^d$ and 
$\ymap:\bbR^{d}\times\bbR^e\!\rightarrow\bbR^e$ are the expected update functions,
\mbox{$\xmds\!\in\bbR^d$} and \mbox{$\ymds\!\in\bbR^e$} are noise terms,
and \mbox{$\xbias\!\in\bbR^d$} and  \mbox{$\ybias\!\in\bbR^e$} are bias terms.
\vspace{-0.7mm}

We make the following assumptions about the noise and bias terms,
which are common in the  stochastic approximation \cite{Borkar08a}.
\vspace{-1mm}

\begin{assumption}
 \label{assum:noise}
 \ifpreprint 
 The stochastic series $\{\mds{i}_n\}$ for \mbox{$i= 1,2$} is a martingale diﬀerence sequence, 
 i.e., with respect to the increasing $\sigma$-ﬁelds,
 \[
 \mathcal{F}_n \triangleq \sigma(\xn[m],\yn[m],\xmds_m,\ymds_m, m\leq n),
 \]
 for some constant $K>0$, the following holds,
 \begin{align}
  \E[\, \mds{i}_{n+1}  \given \mathcal{F}_n ] &= 0, \ \  \forall n \in \{1,2,\dots\},
  \notag\\
  \E[\,\norm{ \mds{i}_{n+1} }^2 \given \calF_n]
  &\leq
  K( 1 + \norm{\xn}^2 + \norm{\yn}^2),
  \ \  \forall n \in \{1,2,\dots\}.
  \notag
 \end{align}
 \else 
  The stochastic series $\{\mds{i}_n\}$ for \mbox{$i= 1,2$} is a martingale diﬀerence sequence, 
 i.e., with respect to the increasing $\sigma$-ﬁelds,
 $\mathcal{F}_n \triangleq \sigma(\xn[m],\yn[m],\xmds_m,\ymds_m, m\leq n)$,
 for some constant $K>0$, the following holds for all $n\in \{1,2,\dots\}$,
 \begin{align}
  \textstyle
  &\E[\, \mds{i}_{n+1}  \given \mathcal{F}_n ] = 0,
  \notag
  \\[-0.5mm] 
  \textstyle
  &\E[\,\norm{ \mds{i}_{n+1} }^2 \given \calF_n]
  \leq
  K( 1 + \norm{\xn}^2 + \norm{\yn}^2).
  \notag
 \end{align}
 \fi 
\end{assumption}
%
\begin{assumption}
 \label{assum:bias}
 \ifpreprint\else\vspace{-1mm}\fi
 The bias $\{\bias{i}_n\}$ for $i=1,2$ is a deterministic or random bounded sequence which is $o(1)$, i.e.,
 \ifpreprint 
 \begin{align}
  \lim_{n\rightarrow\infty} \bias{i}_n = 0.
  \notag
 \end{align}
 \else 
 $\lim_{n\rightarrow\infty} \bias{i}_n = 0$.
 \ifpreprint\else\vspace{-2mm}\fi
 \fi 
\end{assumption}

In our problem setting, the size of the history set $\calH$ is bounded, which ensures that the assumptions regarding the noise and bias terms in Eqs.~\eqref{eq:x_update_n} and \eqref{eq:y_update_n} are realistic and not overly restrictive.

For our analysis, we use the ordinary diﬀerential equation (ODE) approach 
for the stochastic approximation \cite{bertsekas96a,Borkar08a}.
The limiting ODEs that Eqs.~\!\eqref{eq:x_update_n} and \eqref{eq:y_update_n}
might be expected to track asymptotically is, for $\tau\geq0$,
\begin{align}
 \dot\x(\tau) &=  \xmap(\x(\tau),\y(\tau)),
 \label{eq:x_update_t}
 \\
 \dot\y(\tau) &=  \ymap(\x(\tau),\y(\tau)).
 \label{eq:y_update_t}
\end{align}
We also make the assumption about the expected update functions $\xmap$ and $\ymap$.
\vspace{0.2mm}

\begin{assumption}
 \label{assum:update_function}
 \ifpreprint 
 The functions $\xmap$ and $\ymap$ be Lipschitz continuous maps, i.e.,
 for some constants $L_1, L_2 < \infty$,
 \begin{align}
  \norm{\xmap(\x,\y) - \xmap(\x',\y')} &\leq L_1 \norm{[\x,\y]-[\x',\y]},\, \forall (\x,\y,\x',\y'),
  \notag
  \\
  \norm{\ymap(\x,\y) - \ymap(\x',\y')} &\leq L_2 \norm{[\x,\y]-[\x',\y]},\, \forall (\x,\y,\x',\y').
  \notag
 \end{align}
 \else 
 The functions $\xmap$ and $\ymap$ be Lipschitz continuous maps.
 \fi 
\end{assumption}
\vspace{0.2mm}
%
\begin{assumption}
 \label{assum:equilibrium}
 \ifpreprint\else\vspace{-1mm}\fi
 The ODE of Eq.~\!\eqref{eq:y_update_t} has a globally asymptotically stable equilibrium $\varphi(\x)$, 
 where \mbox{$\varphi:\bbR^d\rightarrow\bbR^e$} is a Lipschitz map.
 \ifpreprint\else\vspace{-1mm}\fi
\end{assumption}
\vspace{0.2mm}
\begin{assumption}
 \label{assum:xybound}
  \ifpreprint\else\vspace{-1mm}\fi
 $\sup_n(\norm{\xn}+\norm{\yn})<\infty$.
 \ifpreprint\else\vspace{-1mm}\fi
\end{assumption}
We first show the convergence analysis result for the case of that 
the mixing probability $\mfun$ is a constant or a fixed function
such as $\mfun:\calH\rightarrow[0,1]$.
%
\begin{proposition}
 \label{prop:mfun_invariant}
 Assume Assumptions \ref{assum:noise}--\ref{assum:xybound} hold.
 Let the mixing probability function $\mfun:\calH\rightarrow[0,1]$ be invariant to the number of episodes $n$,
 and the learning rates $\xlr_n$ and $\ylr_n$ satisfying
\ifpreprint 
\begin{align}
 \left\{
 \begin{array}{l}
  \displaystyle \
  \lim_{N\rightarrow\infty} \sum_{n=0}^N \xlr_n
  = \lim_{N\rightarrow\infty} \sum_{n=0}^N \ylr_n
  = \infty,
  \\[5mm]
  \displaystyle \
  \lim_{N\rightarrow\infty} \sum_{n=0}^N \big( \xlr_n^2 + \ylr_n^2 \big) 
  < \infty ,
  \\[5mm]
  \displaystyle \
  \lim_{N\rightarrow\infty}\frac{\xlr_N}{\ylr_N} = 0.
 \end{array}
\right.
\label{eq:two_timescale_learning_rate}
\end{align}
\else 
\begin{align}
 \left\{
 \begin{array}{l}
 \lim_{N\rightarrow\infty} \sum_{n=0}^N \xlr_n
 = \lim_{N\rightarrow\infty} \sum_{n=0}^N \ylr_n
 = \infty,
 \\[2mm]  
 \lim_{N\rightarrow\infty} \sum_{n=0}^N \big( \xlr_n^2 + \ylr_n^2 \big) 
 < \infty,
 \\[2mm]  
 \lim_{N\rightarrow\infty}\frac{\xlr_N}{\ylr_N} = 0.
  \end{array}
 \right.
 \notag
 \\[-12mm]
 \label{eq:two_timescale_learning_rate}
 \\[1mm]
 \notag
\end{align}
\fi 
Then, almost surely, the sequence $\{(\xn,\yn)\}$ generated by Eqs.~\!\eqref{eq:x_update_n} and \eqref{eq:y_update_n}
converges to a compact connected internally chain transitive invariant set of 
Eqs.~\!\eqref{eq:x_update_t} and \eqref{eq:y_update_t}, respectively.
\end{proposition}
%
%
%
The above results indicate how the learning rates $\xlr_n$ and $\ylr_n$ should be set
for convergence.
Since $\ylr_n$ in an MCTL policy is basically proportional to $\frac{1}{n}$,
an obvious choice of $\xlr_n$ will be $\frac{1}{1+n\log n}$.
Note that,
if a deterministic policy such as UCT (Eq.~\!\eqref{eq:uct}) is used, 
$\xmap$ and $\ymap$ will not be the Lipschitz maps and thus the above convergence results cannot be applied.
Therefore, we will use the softmax function \cite{sutton2nd}
in the implementation in Section \ref{subsec:implementation}.

The invariant condition of $\mfun$ can be relaxed.
%
\begin{proposition}
 \label{prop:mfun_adaptive}
 Let $\mfunn[\xn]:\calH\rightarrow[0,1]$ be a function parameterized by a part of $\x$
 and a Lipschitz continuous map with respect to $\x$.
 Assume that all the conditions of Proposition \ref{prop:mfun_invariant} are satisfied 
 except for $\mfun$.
 The consequence of Proposition \ref{prop:mfun_invariant} still holds.
\end{proposition}

Finally, we consider a specific scenario that $\mfunn$ is a decreasing function of the number of episodes $n$, 
where an MCTL policy $\poly$ 
is just used for guiding the PG. 
The goal, in this case, will be to obtain 
a parameterized policy $\polx$ that demonstrates good performance by itself.
In the case of $\mfunn$ decreasing, the convergence condition can be significantly relaxed as follows.
%
\begin{proposition}
\label{ref:mbun_decreasing}
 Assume  Assumptions \ref{assum:noise} and \ref{assum:bias} only for $i=1$ hold,
 $\xmap$ is Lipschitz continuous map, and $\sup_n(\norm{\xn})<\infty$ holds.
 Let the mixing probability $\mfunn$ be $o(1)$ and satisfy $0\leq \mfunn \leq 1-\varepsilon$ for all $n$
 and a constant $\varepsilon>0$, and the learning rate of a PG policy satisfy
\begin{align}
  \lim_{N\rightarrow\infty} \sum_{n=0}^N \xlr_n  = \infty, \quad
  \lim_{N\rightarrow\infty} \sum_{n=0}^N  \xlr_n^2   < \infty.
\notag
\end{align}
Then, 
almost surely, the sequence $\{\xn\}$ generated by Eqs.~\!\eqref{eq:x_update_n} and \eqref{eq:y_update_n}
converges to a compact connected internally chain transitive invariant set of the ODE, $\dot\x(\tau) = \difx \obj(\polxt)$.
\end{proposition}
This proposition shows that, unlike the previous cases,
the convergence property is guaranteed even if a deterministic policy like UCT is used  for an MCTL policy.

\subsection{Implementation}
\label{subsec:implementation}

We present an implementation of PG-MCTL that satisfies the convergence conditions
and converges to a reasonable solution.

First, we consider revising the update of a PG policy $\polx$.
Since the goal is to maximize the expected return of Eq.~\!\eqref{eq:obj}, 
the following update at each episode $n$ will be appropriate, 
instead of the ordinary one of Eq.~\!\eqref{eq:pg_update}:
\ifpreprint
\begin{align}
 \xn[n+1] 
 &
 =\xn+ \xlr_n \sum_{t=0}^\tmax \difx\log \polxyn(a_t|h_t) (g_t - b(h_t))
 \notag
 \\[-0mm]
 &
 = \xn + \xlr_n \sum_{t=0}^\tmax \rho_t  \difx\log \polxn(a_t|h_t) (g_t - b(h_t)),
 \label{eq:pg_update_mixpolicy}
\end{align}
\else
\begin{align}
 \xn[n+1] 
 &=\xn+ \xlr_n \sum_{t=0}^\tmax \difx\log \polxyn(a_t|h_t) (g_t - b(h_t))
 \notag
 \\[-0mm]
 &
 = \xn + \xlr_n \sum_{t=0}^\tmax \rho_t  \difx\log \polxn(a_t|h_t) (g_t - b(h_t)),
 \label{eq:pg_update_mixpolicy}
\end{align}
\fi
where we assume $\polxyn$ is the behavior policy of episode $n$.
The $\rho_t$ is a scaled probability ratio or 
a kind of importance weight \cite{sutton2nd},
\begin{align}
 \rho_t \triangleq  \frac{(1-\mfun(h_t))\polx(a_t|h_t)}{\polxy(a_t|h_t)}.
 \label{eq:iw}
\end{align}
%

Note that according to Proposition \ref{prop:mfun_adaptive},
learning the mixture probability $\mfun$ via this PG update,
rather than assuming $\mfun$ is predefined and fixed,
still ensures convergence.
When adopting this approach, termed \textbf{PG-MCTL-adpt},
we assume that $\mfunn[\x]$ is parameterized by a subset of parameters within $\x$.
The update rule of $\x$ is derived as
\ifpreprint
\begin{align}
 \xn[n+1] = \xn+ \xlr_n \sum_{t=0}^\tmax
 \bigg(
 \rho_t \difx\log \polxn + \frac{(\polyn - \polxn)}{\polxyn} \difx \mfunn[\xn]
 \bigg)
 (g_t - b(h_t)).
 \label{eq:pg_update_mixpolicy_lambda}
\end{align}
\else
\begin{align}
 \hspace{-2mm}
 \xn[n+1] = \xn +
 &
 \xlr_n \sum_{t=0}^\tmax\!
 \bigg\{
 \rho_t \difx\log \polxn(a_t|h_t) + \frac{(\polyn(a_t|h_t) - \polxn(a_t|h_t))}{\polxyn(a_t|h_t)} \difx \mfunn[\xn]\!(h_t)
 \!\bigg\}
 (g_t - b(h_t)).
 \label{eq:pg_update_mixpolicy_lambda}
\end{align}
\fi

From here on, we will only consider the case where the mixing probability $\mfun$ is constant, 
but the results presented here will be straightforwardly applied to other settings of $\mfun$.
\vspace{-0.6mm}

Next, we consider an implementation of an MCTL policy $\poly$.
The update rule of MCTS that MCTL follows differs from the general
stochastic approximation in Eq.~\!\eqref{eq:y_update_n}.
In particular, the learning rate in the MCTS update of Eq.~\!\eqref{eq:mcts_update}
varies among nodes.
%
%
On the other hand, the stochastic approximation assumes
a global learning rate $\ylr_n$ as seen in Eq.~\eqref{eq:y_update_n}.
Furthermore, although Assumption \ref{assum:xybound}, vital for convergence, 
states that parameters should remain bounded,
the parameter $m$, the number of visits, could diverge.
To reconcile these differences, we introduce a tree-inclusion probability and reformulate the MCTS update
by replacing $m(h,a)$ with $\invvst\!:\!\calH\!\times\!\calA\rightarrow[0,1]$
so that \mbox{$m(h,a)\!=\!1/\invvst(h,a)$}.
Consequently, the parameter of the MCTL policy $\poly$ becomes $\y\triangleq\{\invvst, \qhat\}$.
With a learning rate of $\ylr_{n}=1/n$ and initializing $\invvst\!:=\!1$ and $\qhat\!:=\!0$, 
the conventional MCTS update in Eq.~\!\eqref{eq:mcts_update} can be rewritten as:
\ifpreprint
\begin{align}
 \begin{cases}
  \displaystyle
  \invvst_{n+1}(h_t,a_t) 
  &\!\!\!\! \displaystyle
  = \invvst_n(h_t,a_t) + \ylr_{n} \kappa_{n,t} \frac{-\invvst_n(h_t,a_t)}{\invvst_n(h_t,a_t) + 1},
  \\[2mm]
  \qhat_{n+1}(h_t, a_t)
  &\!\!\!\! \displaystyle
  = \qhat_n(h_t,a_t) + \ylr_n \kappa_{n,t} ( \ret_t - \qhat_n(h_t,a_t)),
  \end{cases}
 \label{eq:mcts_update_2}
\end{align}
\else\vspace{0mm}
\begin{align}
 \begin{cases}
  \invvst_{n+1\!}(h_t,a_t) 
  &\!\!\!\!\!\! 
  = \invvst_{n\!}(h_t,a_t) + \ylr_{n} \kappa_{n,t} \frac{-\invvst_n(h_t,a_t)}{\invvst_n(h_t,a_t) + 1},
  \\[2.5mm]
  \qhat_{n+1\!}(h_t, a_t)
  &\!\!\!\!\!\! 
  = \qhat_{n\!}(h_t,a_t) + \ylr_n \kappa_{n,t} ( \ret_t - \qhat_n(h_t,a_t)),
  \end{cases}
 \label{eq:mcts_update_2}
\end{align}
\fi
where $\kappa_{n,t}$ is the following and can be regarded as an adjustment term for the learning rate per node,
\begin{align}
 \kappa_{n,t} &\triangleq  p_{n,t}\, \frac{\invvst_n(h_t,a_t)}{\ylr_n},
 \notag
\end{align}
%
and $p_{n,t}$ is the following tree-inclusion probability,
\begin{align}
 p_{n,t} &\triangleq
 \begin{cases}
  1,
  & \textrm{if\ } t=0,
  \\
  \min\!\left( \frac{1}{\invvst_n(h_{t-1},a_{t-1})} - 1, 1 \right),
  & \textrm{otherwise}.
 \end{cases}
 \label{eq:treeIncProb}
\end{align} 
If $\invvst_n(h_{t-1},a_{t-1})=1$, the tree-inclusion probability $p_{n,t}$ is zero,
and thus the values of $(h_t,a_t)$ are not updated.
It corresponds to the case 
where the tree does not have a node $(h_t,a_t)$, indicating that this node hasn't been expanded.
If $\invvst_n(h_{t-1},a_{t-1})\leq0.5$, the values of a node $(h_t,a_t)$ are  updated with probability $1$.
The equivalence of the original MCTS update and Eq.~\eqref{eq:mcts_update_2} is 
shown in Appendix~\ref{appendix:equivalence}.
%

We next investigate Assumption \ref{assum:update_function} about Lipschitz continuity
of the expected update functions $\xmap$ and $\ymap$.
The PG update of Eq.~\eqref{eq:pg_update_mixpolicy} is based on the gradient ascent,
and thus the expected update functions $\xmap$ with an ordinary implementation will satisfy Lipschitz continuity.
However, the  update of Eq.~\eqref{eq:mcts_update_2} does not allow $\ymap$ to have Lipschitz continuity 
since $\kappa_{n,t}$ diverges as $\ylr_n \rightarrow 0$.
This problem can be solved by modifying $\kappa_{n,t}$ with a large value $M>0$ as follows:
\begin{align}
 \bar\kappa_{n,t} &\triangleq \min(\kappa_{n,t},M).
 \label{eq:bounded_adjustment}
\end{align}

\begin{theorem}
 \label{theo:convergence}
 Let the PG-MCTL update the parameterized policy $\polx$ by Eq.~\!\eqref{eq:pg_update_mixpolicy}
 and the MCTL policy $\poly$ by the rule in which
 $\kappa$ in Eq.~\!\eqref{eq:mcts_update_2} is replaced by $\bar\kappa$ of Eq.~\!\eqref{eq:bounded_adjustment},
 and the learning rates satisfy the conditions of Eq.~\!\eqref{eq:two_timescale_learning_rate}.
 Also let $\polx$ be defined on a compact parameter space and have always bounded first and second partial derivatives, 
 and $\poly$ be a softmax policy with hyper-parameters $\beta\geq 0$ and $C\geq0$,
 \ifpreprint 
 \begin{align}
  \poly\!(a|h) \propto
  \exp\!\!\left(\!\!
  \beta\Big\{\qhat(h,\!a) + C\!\sqrt{ \invvst(h,\!a) \log\! \sum_b \frac{1}{\invvst(h,\!b)}} \Big\}
  \!\!\right)\!\!.
  \label{eq:softUCT}
 \end{align}
 \else 
 \vspace{-1.5mm}
 \begin{align}
  \textstyle
  \poly\!(a|h) \propto
  \exp\left(
  \beta\Big\{\qhat(h, a) + C \sqrt{\invvst(h, a) \log \sum_b \frac{1}{\invvst(h, b)}} \Big\}
  \right).
  \label{eq:softUCT}
 \end{align} 
\fi
 Then, 
 $\lim_{n\rightarrow\infty} \difx \obj(\polxyn)=0$ holds.
\end{theorem}
%
%

Finally, we propose a heuristic to avoid a vanishing gradient problem of the PG update of Eq.~\eqref{eq:pg_update_mixpolicy}.
By the definition of $\rho_t$ in Eq.~\!\eqref{eq:iw}, if $\polx$ and $\polxy$ are significantly different,
$\rho_t$ can be close to zero and thus the stochastic gradient at time $t$ can vanish.
In order to avoid this problem,
we modify $\rho_t$ to $\underline{\rho}_t$ as, with $\upsilon\in[0,1]$,
\ifpreprint\else\vspace{-2mm}\fi
\begin{align}
 \underline{\rho}_t \triangleq \max(\upsilon, \rho_t).
 \label{eq:bounded_rho}
 \ifpreprint\else\vspace{-1mm}\fi
\end{align}
When $\upsilon_n=o(1)$, the convergence property in Theorem \ref{theo:convergence} still holds
because $\upsilon_n$ is absorbed into $\epsilon_n$ in Eq.~\!\eqref{eq:x_update_n}.
Note also that $\rho_t$ is upper bounded by 1. 
Thus there is no need to care about $\rho_t$ taking a large value.
\ifpreprint 

The entire procedure of this PG-MCTL implementation is shown in Algorithm \ref{algo}.

\input{pg_mcts_algo.tex}

\else 
The entire PG-MCTL implementation is shown in Algorithm \ref{algo}. 

\input{pg_mcts_algo.tex}
\fi 

It should be noted that the memory size may continue to grow over time as more episodes are experienced.
Additionally, the maximum memory usage of the MCTL policy is on the order of the tree size, $\mathcal{O}(|\calH|)$. 
While this can be significant, it is generally not expected to explore all trajectories and construct a complete tree, resulting in a smaller memory consumption.

%% file: pg_mcts_algo.tex
\newcommand{\STATE}{\State}
\newcommand{\FOR}[1]{\For{#1}}
\newcommand{\ENDFOR}{\EndFor}
\newcommand{\WHILE}[1]{\While{#1}}
\newcommand{\ENDWHILE}{\EndWhile}
\newcommand{\RETURN}[1]{\State \Return{#1}}

\begin{algorithm}[ht]
 \caption{A PG-MCTS implementation}
 \label{algo}
 \begin{algorithmic}[1]
 \newcommand{\q}{\hspace*{0.6em}} \newcommand{\qq}{\hspace*{1.2em}}
 \newcommand{\pp}{\hspace*{-1.0em}} \newcommand{\p}{\hspace*{-0.6em}}
  \STATE \textbf{given:} 
  \STATE \ \ - an initialized PG policy $\polx(a|h)$ and mixing probability function $\mfunn[\x](h)$ \footnotemark
  \STATE \ \ - an initialized MCTL policy $\poly(a|h)$, e.g., Eq.~\!\eqref{eq:softUCT} 
  \STATE \ \ - hyper-parameters for the PG and MCTS policies
  \WHILE{within computational budget}
  \STATE {\texttt{\color{blue}// interaction with environment $\Env$}}
  \STATE observe an initial observation $h_0 \sim \pini$
  \STATE empty a memory $\mathcal{M}$ and store $h_0$ in $\mathcal{M}$ 
  \FOR{$t=0$ {\bfseries to} $\tmax$}
  \STATE choose a policy $\pi\in\{\polx,\poly\}$, using $\lambda(h_t)$ (see Eq.~\!\eqref{eq:polxy})
  \STATE choose and execute an action $a_t \sim \pi(\,\cdot\,|h_t)$
  \STATE observe a reward $r_{t} := \rfun(h_t, a_t)$
  \STATE observe a new history $h_{t+1}\sim\ph(\,\cdot\,|h_t,a_t)$
  \STATE store $r_t$ and $h_{t+1}$ in the memory $\mathcal{M}$
  \ENDFOR
  \STATE \texttt{\color{blue}// update of policies $\polx,\,\poly$ and mixing probability $\mfunn[\x]$ with $M$}
  \STATE compute the return $g_t,\ \forall t\in\{0,\dots,\tmax\}$
  \STATE update  $\polx$ and $\mfunn[\x]$ by the PG update of Eqs.~\!\eqref{eq:pg_update_mixpolicy} and \eqref{eq:pg_update_mixpolicy_lambda}  with \eqref{eq:bounded_rho}
  \STATE update $\poly$ by the MCTL update of Eq.~\!\eqref{eq:mcts_update_2} with \eqref{eq:bounded_adjustment}
  \ENDWHILE
  \RETURN the learned policy 
   $\polxy \triangleq \mfun\polx + (1-\mfun)\poly$
 \end{algorithmic}
\end{algorithm}
\footnotetext{The mixing probability function $\mfun$ may be given as a hyper-parameter instead of parameterizing it with $\x$. In that case, the update $\mfun$ of line 18 by Eq.~\eqref{eq:pg_update_mixpolicy_lambda} is skipped.}

%% file: sec_related_work.tex
\section{Related work}
\label{sec:related_work}

\ifpreprint 
There are a lot of studies that integrate MCTS and RL algorithms
\cite{Guo14a,Vodopivec17a,Silver17a,Jiang18a,Efroni19a,Ma19a,Schrittwieser20a,Grill20a,Dam21a}.
\else 
Numerous studies integrate MCTS and RL algorithms.
\fi 
Most of them are based on the standard MCTS setting (decision time planning)
and propose to use value-based RL \cite{Vodopivec17a,Jiang18a,Efroni19a} 
or supervised learning \cite{Guo14a,Silver17a,Anthony17a,Schrittwieser20a,Dam21a}, 
where deep neural networks are trained from targets generated by the MCTS iterations. 
The latter approach is also known as expert iteration \cite{Anthony17a}.
AlphaZero and MuZero are prominent algorithms adopting this approach. 
The key distinction between expert iteration and PG-MCTL lies in their policy updates.
While the PG-MCTL is weighting the experiences with the return $\ret_t$ and importance weight $\rho_t$ (see Eq.~\eqref{eq:pg_update_mixpolicy}),
the standard expert iteration does not, assuming that all instances are positive examples since they are the result of MCTS iteration.
%
%
%
Another difference pertains to the type of learning. 
Specifically, the expert iteration is classified as decision-time planning or model-based RL.
In contrast, PG-MCTL is model-free and doesn't rely on a simulator.
\citet{Grill20a} also extend AlphaZero or MuZero with the notion of PG, which is also model-based RL.
\vspace{-0.5mm}

Several studies, among others, combine PG and MCTS%
\ifpreprint \cite{Guo16a,Anthony19a,Soemers19a,Dieb20a}.
\else. \fi
%
\citet{Soemers19a} runs MCTS to compute a value function that PG uses.
%
\citet{Guo16a} uses PG to design reward-bonus functions to improve the performance of MCTS.
%
\citet{Anthony19a} 
uses PG for updating local policies
and investigates planning without an explicit tree search.
%
\citet{Dieb20a} uses  PG in the tree expansion phase to choose a promising child node to be created,
assuming a situation where the number of actions is huge. 
\vspace{-0.5mm}

From another perspective, the PG-MCTL can be regarded 
as using MCTS for PG to enhance the efficiency of exploration.
%
Most exploration approaches in PG focus on designing reward functions, 
often incorporating a bonus of intrinsic motivation or curiosity to explore unknown states
\cite{Bellemare16a,Tang17a,Zheng18a,Burda19a}.
\citet{Haarnoja18a} demonstrates remarkable success using an entropy bonus to aid exploration in benchmark control tasks.
%
Unlike many approaches, PG-MCTL does not modify the objective function,
however, it remains compatible with most of them.
\vspace{-0.5mm}

For RL in an HDP or NMDP, there are two major directions.
The first one assumes the existence of latent dynamics and considers the identification of the dynamics
\ifpreprint
\cite{Thiebaux06a,Poupart08a,Silver10a,Singh12a,DoshiVelez15a,Brafman19a}.
\else
\cite{Thiebaux06a,Poupart08a,Silver10a,Singh12a,DoshiVelez15a,Brafman19a}.
\fi
%
A POMDP \cite{Kaelbling98a} serves as a widely-recognized mathematical model for this purpose.
%
\citet{DoshiVelez15a} identifies an environment as a POMDP 
with Bayesian non-parametric methods 
and then compute a policy by solving the POMDP. 
The other direction is to use a function approximator 
whose output depends not only on a current observation 
but also on past observations \cite{Loch98a,HernandezGardio00a,Bakker02a,Hausknecht15a,SCST,Qin23a}.
One of the successful approaches uses a neural network for sequence learning 
as a policy model and optimizes it by PG \cite{Wierstra10a,SCST,Paulus18a,Kamigaito21a}, 
as corresponds to the PG policy in our proposed PG-MCTL.
\vspace{-0.5mm}

While the proposed implementation of PG-MCTL integrates standard PG and MCTS algorithms in a well-designed way,
there are a lot of studies on enhancing those algorithms, 
such as stabilization of PG by a conservative update \cite{kakade02a,Schulman15a,Schulman17a},
the entropy regularization for explicitly controlling the exploration-exploitation trade-off \cite{Haarnoja17a,Haarnoja18a,Xiao19a,Grill20a},
and extensions of MCTS to continuous spaces \cite{Couetoux11a,Mansley11a,Kim20a,Mao20a}.
Incorporating these technologies, including the expert iteration, into the PG-MCTL is an interesting avenue for future work.

%% file: sec_experiments.tex
\section{Numerical Experiments}
\label{sec:experiments}

We apply the PG-MCTL algorithm to two different tasks in HDPs.
The first task is a randomly synthesizing task,
which does not contain domain-specific structures and is not overly complex.
Therefore, this task will help investigate the primary performance of algorithms.
The second task is the long-term dependency T-maze, which is known as
a standard benchmark for learning a deep-memory POMDP \cite{Bakker02a,Wierstra10a}.
%
%
Details of the experimental setup are given in Appendix \ref{appendix:exp_setup}.

The goal here is not to ﬁnd the best model for the above two tasks,
but to investigate if/how combining the PG and MCTL (MCTS variant) by the PG-MCTL is effective.
%
Therefore, the applied algorithms here are simple, 
not state-of-the-art algorithms. 
In this regard, model-based RLs including MuZero \cite{Schrittwieser20a} are also out of the scope of this work.
We used REINFORCE with a baseline \cite{williams92a} for the PG 
and MCTL for the original MCTS, which are introduced
in Sections \ref{subsec:pg} and \ref{subsec:approach}, respectively.
Note that REINFORCE, although it is classic, is still appealing 
due to its good empirical performance and simplicity \cite{Grooten22a,Zhang21a}.
It and its variants are used in many applications
\cite{SCST,Paulus18a,Chen19a,Xia20a,Wang21a,Liu22a}.
Thus, we believe that improving REINFORCE itself is still important 
in the practical implementation of RL.
%

While our focus is on fundamental algorithms, we also included the proximal policy optimization (PPO) \cite{Schulman17a}, a modern and practical variation of the PG algorithm, to provide a comparative perspective on performance.
Additionally, we utilized a simple version of AlphaZero \cite{Silver17a}, termed lazy AlphaZero.
It employs the same adaptation to the online RL setting as MCTL (also called lazy MCTS).
The parameterized policy model, serving as the prior policy in lazy AlphaZero,
is updated based on online experiences according to likelihood maximization.
Furthermore, we also implemented a naive mixture of the PG and MCTL that follows Eq.~\eqref{eq:polxy}
but uses the learning rules of the standalone REINFORCE and MCTL. 
%
%
For fair evaluation, we first tuned the hyper-parameters of standalone algorithms
and then used them for the PG-MCTL and the naive mixture model.

\subsection{Randomly synthesized task}
\label{subsec:synthesized_task}
\ifpreprint 
The first task is a randomly synthesized non-Markovian task. 
It is modeled as a simple but challenging HDP 
analogous to the generation tasks such as text generation and compound synthesis.
\else 
This task is a randomly synthesized non-Markovian model.
It is analogous to generation tasks such as text generation and compound synthesis, presenting as a simple yet challenging HDP.
\fi 
There are five observations and ten actions.
The observation probability function $\po$ 
was synthesized to depend on the time-step, observation, and action.
The reward function $\rfun$ was composed of 
the sum of the per-step sub-reward function $r_{\rm local}$ 
and the history-based sub-reward function $r_{\rm global}$.
The function $r_{\rm global}$ was synthesized by using a Gaussian process, 
such that the more similar the histories, the closer their rewards tend to be.
This reward function $\rfun$ can be interpreted in the context of text generation as follows:
$r_{\rm local}$ represents the quality of local word connections, 
and $r_{\rm global}$ represents the quality of the generated text.
The policy $\polx$ was a softmax and parameterized by using the reward structure.
We set $\tmax\!=\!15$.
Thus, there are an enormous number of variations in the histories ($\sim 10^{25}$).
%


Figure \ref{fig:results_synthesized_task} (a) shows the results of ten independent runs,
where 'PG-MCTL' and 'PG-MCTL-adpt' are the proposed methods.
PG-MCTL uses a fixed mixing probability $\mfunn$,
while PG-MCTL-adpt learns $\mfunn[\x]$ with the PG update of \eqref{eq:pg_update_mixpolicy_lambda}.
In each run, an HDP environment was independently generated, as described above.

The results indicate that
REINFORCE learned most quickly but often fell into sub-optimal policies.
In contrast, the MCTL and lazy AlphaZero methods continued to improve but was slow to learn. 
This slowness is probably due to the lack of the ability to learn state representation,
more concretely, if trajectories are even slightly different,
the nodes differ from each other, and thus information is not shared among them.
Whereas, the proposed PG-MCTL methods were able to continuously improve  and was not slow to learn.
It implies that the PG-MCTL approach can successfully incorporate the advantages of both the PG and MCTL.
Specifically, while PG offers rapid learning, MCTL provides continuous improvement without easily falling into sub-optimal policies.
%

However, it is  worth noting that the poor performance of the naive mixture indicates that the simple
mixing approach of the PG and MCTL policies cannot work.
Also, note that their performances of PG-MCTL and PG-MCTL-adpt were similar. 
This similarity suggests that in this task, 
learning the mixing probability $\lambda$ 
did not provide a significant advantage over the fixed probability approach in PG-MCTL.
However, it is important to consider the potential benefits of adaptive methods in more structured tasks, 
as explored in Section~\ref{subsec:tmaze_task}.


\begin{figure*}[t]
 \begin{tabular}{lll}
  \hspace{-5mm} 
  (a)
  &
      \hspace{-1mm} 
      (b) 
      &\hspace{-1mm}  
	  \\
  \hspace{-9mm}
  \includegraphics[width=2.05in]{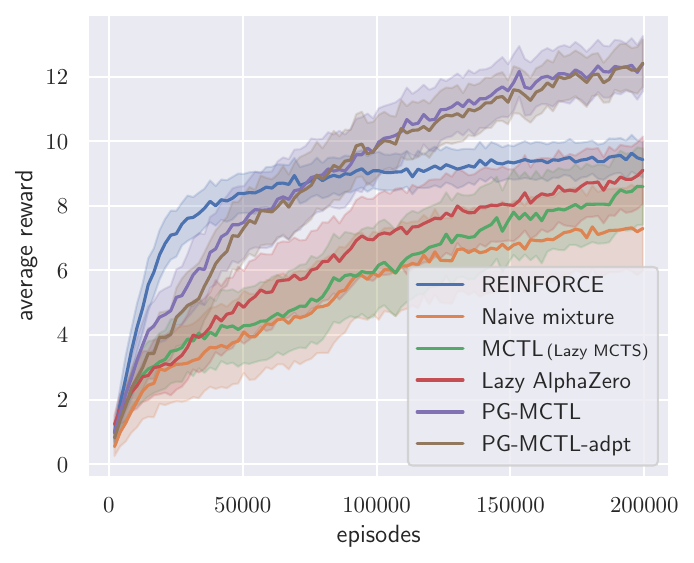}
  &
      \hspace{-4mm}
      \includegraphics[width=2.1in]{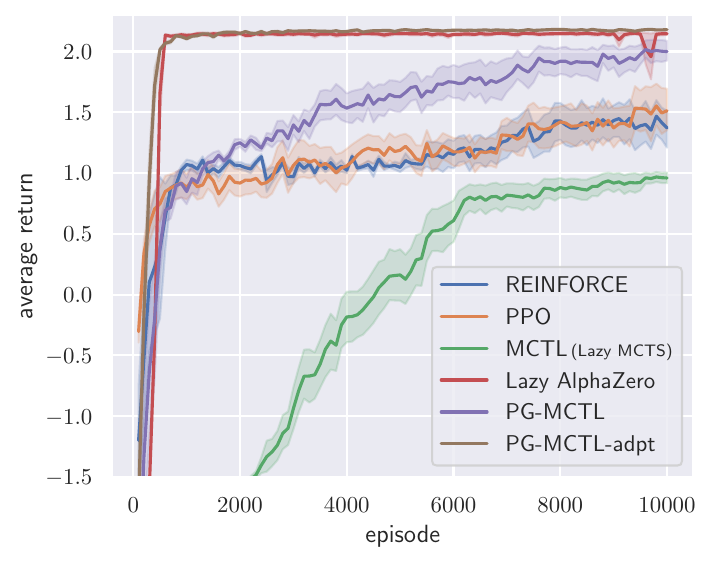}
      &\hspace{-5mm}
	  \includegraphics[width=2.1in]{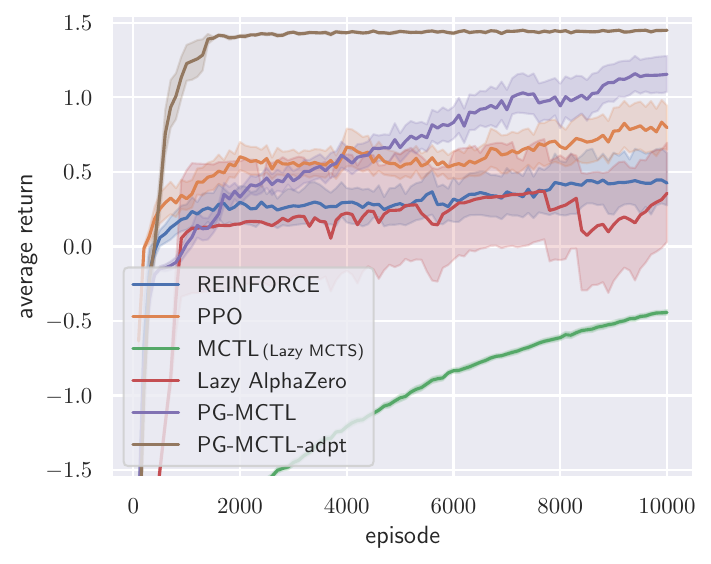}
 \end{tabular}
 \caption{Performance comparison by ten independent runs,
 where the error bar represents the standard error of the mean:
 (a) the randomly synthesized task ($\tmax=15$). 
 (b) T-maze task; 
 the plot on the left is the result of an easy setting (the length of corridor $L = 30$ and the initial position $s_0=0$). The plot on the right is for a more difficult setting, where there exist more sub-optimal policies (the length of corridor $L = 100$ and the initial position $s_0=50$).
}
 \label{fig:results_synthesized_task}
\end{figure*}

\ifpreprint
\begin{figure*}[ht]
 \begin{tabular}{lll}
  \hspace{-3mm} (a)
  &\hspace{-3mm} (b) 
      &\hspace{-1mm}  \\[-0.5mm]
  \hspace{-4mm}\includegraphics[width=2.25in]{figs/random_task_15.pdf}
  &\hspace{-4mm}
      \includegraphics[width=2.3in]{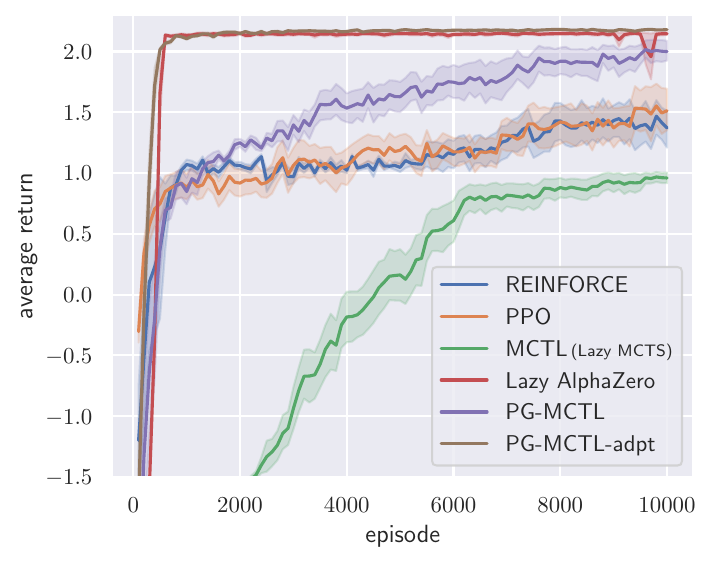}
      &\hspace{-4mm}
	  \includegraphics[width=2.3in]{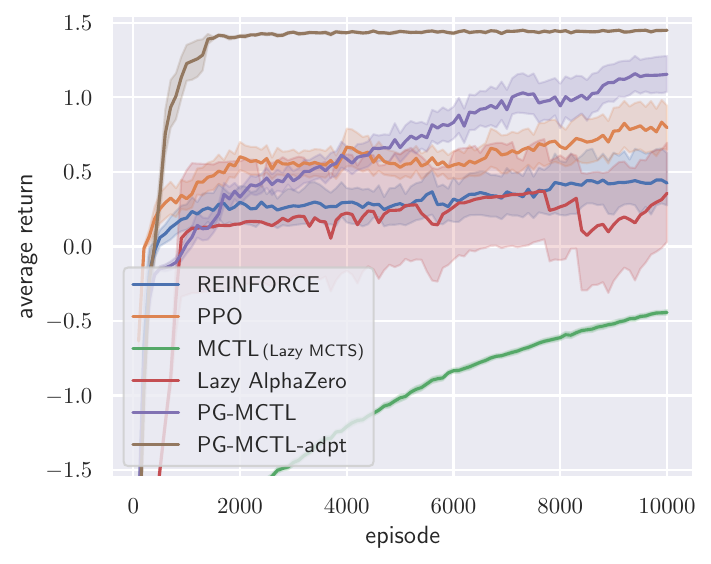}
 \end{tabular}
 \caption{Performance comparison by ten independent runs,
 where the error bar represents the standard error of the mean:
 (a) the randomly synthesized task ($\tmax=15$). 
 (b) T-maze task; 
 the plot on the left is the result of an easy setting (the length of corridor $L = 30$ and the initial position $s_0=0$). The plot on the right is for a more difficult setting, where there exist more sub-optimal policies (the length of corridor $L = 100$ and the initial position $s_0=50$).
}
 \label{fig:results_synthesized_task}
\end{figure*}
\fi

\subsection{T-maze task}
\label{subsec:tmaze_task}
\ifpreprint
The second experiment is the non-Markovian T-maze task \cite{Bakker02a,Wierstra10a} 
(see Figure \ref{fig:tmaze_task}).
It is designed to test the ability to identify associations between events with long-time lags.
\else
The T-maze task of Figure \ref{fig:tmaze_task}
 is non-Markovian and designed to test the ability to identify associations between events with long-time lags \cite{Bakker02a,Wierstra10a}.
\fi
An agent has to remember an observation made at the first time step until the episode ends.
We use a long short-term memory (LSTM) as a policy model.
%
%
\ifpreprint

\begin{SCfigure}[45][h]
 \centering
 \vspace{-0mm}
 \caption{Long-term dependency 
 T-maze task: an agent starts at the position $\textsf S$.
 Only at the initial time step $t=0$, 
 it can observe a signal 'up' or 'down' that indicates 
 it should go north or south at the T-junction in this episode.
 In this example, the direction is 'up',
 the starting position is $s_0=0$,
 and the length of corridor is $L=10$.
 }
 \includegraphics[width=2in]{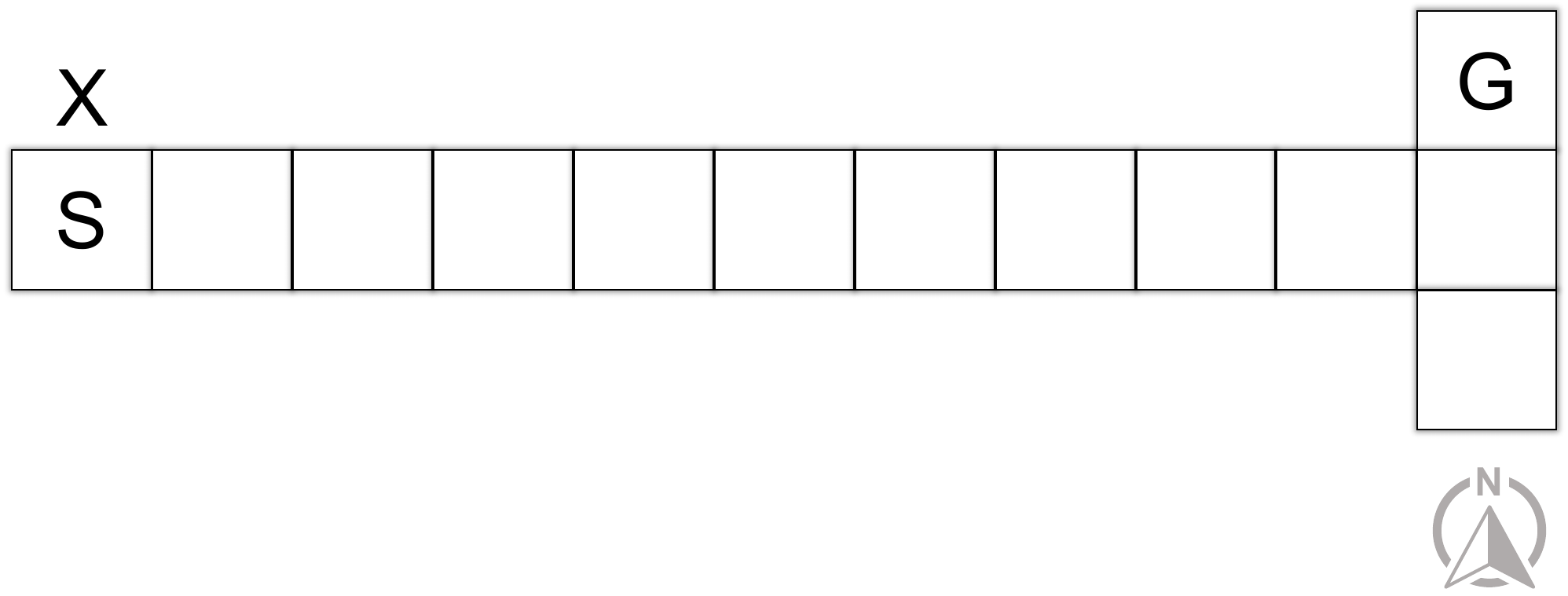}
 \label{fig:tmaze_task}
\end{SCfigure}
\else
\begin{figure}[t]
 \centering
 \vspace{3mm}
 \includegraphics[width=2.35in]{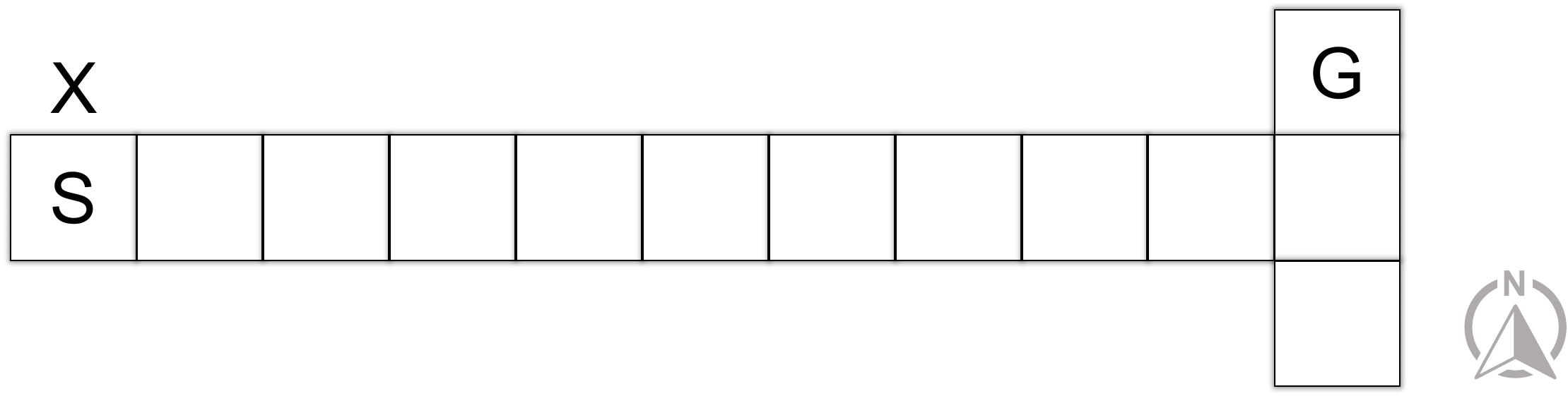}
\vspace{1mm}
 \caption{Long-term dependency 
 T-maze task: an agent starts at the position $\textsf S$.
 Only at the initial time step $t=0$, 
 it can observe a signal 'up' or 'down' that indicates 
 it should go north or south at the T-junction in this episode.
 }
 \label{fig:tmaze_task}
\end{figure}
\fi
Since the original setting with the initial position $s_0 = 0$ is not so difficult,
we prepared a more difficult setting, where initial position $s_0$ is the center of the corridor.
In this setting, a policy that chooses the left or right action with probability $0.5$ 
can be sub-optimal.
Note that this is not true in the original setting ($s_0=0$) 
because going left occurs a negative reward.
\vspace{-0.5mm}

The results are shown in Figure \ref{fig:results_synthesized_task} (b)
and indicate the effectivity of the proposed methods.
In this experiment,
the performance of PPO and REINFORCE showed only marginal differences, 
suggesting that using PPO as the PG module in PG-MCTL would offer limited benefits.
The significant superiority of our approach over PPO, 
however, underscores the effectiveness of combining PG with MCTS. 
Moreover, PG-MCTL-adpt outperformed PG-MCTL,
demonstrating that adjusting the mixing probability through learning, 
especially at T-junctions, was effective in this task.

%% file: sec_conclusion.tex
\section{Conclusion}
\label{sec:conclusion}
\ifpreprint 
In this paper, we focused on history-based decision processes (HDPs) and
investigated the PG-MCTL approach, an approach for mixture policies of the PG and online MCTS variant (MCTL),
in order to take advantages of the characteristics of both the PG and MCTS frameworks. 
We provided the convergence analysis and then proposed an implementation that converges to a reasonable solution.  
Through the numerical experiments with two simple HDP tasks, 
we confirmed significant effect of the proposed approach for the mixture of the PG and MCTL policies.
To the best of our knowledge, this is the first study on policy gradient algorithms guided by the MCTS framework
with theoretical justification,
though there exist studies of MCTS algorithms using the PG to solve partial problems in the MCTS.

In the future work, we will apply our algorithms with the state-of-the-art neural networks for sequence data
to more practical and challenging domains, 
such as text summarization, advertising text generation, and incomplete information games.
Further theoretical analysis, especially convergence rate analysis,
is necessary to more deeply understand the properties of the PG-MCTS
and to improve the algorithms.
Also, the analysis of convergence points is interesting 
because the PG is a local optimization while MCTS is a global optimization method.
For example,  a guide by MCTS may help PGs get out of a bad local optima as well as a learning plateau.
%
Incorporating the state-of-the-art techniques in the PG and MCTS, such as 
the entropy regularization and the natural gradient, will be another interesting direction.
%
\else 
This paper focused on online reinforcement learning problems in history-based decision processes (HDPs). 
We investigated the PG-MCTL approach, a mixture policy approach for the PG and online MCTS variant (MCTL) that takes advantage of the features of the PG and MCTS algorithms. 
We provided the convergence analysis and then proposed an implementation that converges to a reasonable solution.  
Through the numerical experiments on two HDP tasks with different characteristics, we confirmed the significant effect of the proposed approach for the mixture of the PG and MCTL policies.
%

In future work, we will apply our algorithms with state-of-the-art neural networks for sequence data to more practical and challenging domains, 
such as advertising text generation and incomplete information games.
Also, the analysis of convergence points is crucial because the PG is a local optimization while MCTS is a global optimization method.
For example, guidance from MCTS may help PGs overcome a bad local optimum and a learning plateau.
Another exciting direction will incorporate state-of-the-art PG and MCTS techniques, such as entropy regularization and the natural gradient.
%
\fi 

%% file: supplement.tex

\ifpreprint\else
\fi

\section{Proofs}
\label{appendix:proofs}
\subsection{Preliminaries}

We first introduce the basic results of the ordinary differential equation (ODE) based approach 
for the stochastic approximation \cite{Borkar08a}.
We consider the following update rule of $\x\in\bbR^d$ with an initial value $\z_0\in\bbR^d$
for all $n\in[0,1,\dots] = \bbN_{\geq0}$,
\begin{align}
 \zn[n+1] &= \zn + \zlr_n [\, \zmap(\zn) + \zmds_{n+1} + \zbias_n].
 \label{eq:z_update_n}
\end{align}
To take the ODE approach,
we extend the above discrete-time stochastic process of $\z_n$
to a continuous, piecewise-linear counterpart $\bar\z:\bbR_{\geq0}\rightarrow\bbR^d$ as follows:
Define a time-instant function $t: \bbN_{\geq0}\rightarrow\bbR_{\geq0}$ 
such as
\begin{align}
 t(n) \triangleq
 \begin{cases}
  0 & \textrm{if\ \ } n=0,
  \\
  \sum_{m=0}^{n-1} \zlr_m  & \textrm{otherwise},
 \end{cases}
 \notag
\end{align}
and set $\bar\z(t(n)) := \z_n, \ \forall n \in \bbN_{\geq0}$.
Then, for any $n\in\bbN_{\geq0}$, we derive the following linear interpolation,
\begin{align}
 \bar\z(\tau) \triangleq \zn + (\zn[n+1]-\zn)\frac{\tau-t(n)}{t(n+1)-t(n)},\quad  \tau \in I_n,
\end{align}
where $I_n \triangleq [t(n), t(n+1)]$.
As we will show later, 
the key result of the ODE approach to the analysis of Eq.~\eqref{eq:z_update_n}
is that $\bar\z(\tau)$ asymptotically almost surely approaches the solution set of the following ODE,
\begin{align}
 \dot\z(\tau) = \zmap(\z(\tau)),\quad \tau\in \bbR_{\geq0}.
 \label{eq:z_ode}
\end{align}

For this purpose, we need to make the following assumptions.

\begin{assumption}
 \label{append_assum:learning_rate}
 The learning rates $\{\alpha_n\}$ are positive scalars satisfying
 \begin{align}
  \lim_{N\rightarrow\infty}\sum_{n=0}^N \alpha_n = \infty,
  \qquad 
  \lim_{N\rightarrow\infty} \sum_{n=0}^N \alpha_n^2 \leq \infty.
 \end{align}
 \\[-5mm]
\end{assumption}

\begin{assumption}
 \label{append_assum:lipschitz}
 The function $\zmap:\bbR^d \rightarrow \bbR^d$ is a Lipschitz continuous map,
 i.e., for some constant $0<L<\infty$,
 \[
 \norm{\zmap(\z)-\zmap(\z')} \leq L \norm{\z-\z'},\qquad  \forall (\z,\z') \in \bbR^d\times\bbR^d.
 \]
 \\[-8.5mm]
\end{assumption}

\begin{assumption}
 \label{append_assum:mds}
 The stochastic series $\{\zmds_n\}$ is a martingale difference sequence
 with respect to the increasing family of $\sigma$-fields
 \[
 \calF_n \triangleq \sigma(\z_i, \zmds_i, \zbias_i, i\leq n).
 \]
 That is, the following holds,
\begin{align}
 \E[\zmds_{n+1} \given \calF_n] &= 0\quad a.s., \quad \forall n\geq0.
 \notag
\end{align}
Furthermore, $\zmds_n$ is always square-integrable with
 \begin{align}
 \E[\, \norm{\zmds_{n+1}}^2 \given \calF_n] &= K(1+\norm{\xn}^2) \quad  \textrm{a.s.},
 \quad \forall n\geq0,
  \label{eq:z_update_t}
\end{align}
for some constant $K\geq0$.
\\[-2.5mm]
\end{assumption}

\begin{assumption}
 \label{append_assum:bias}
 The series of bias $\{\zbias_n\}$ is a deterministic or random bounded sequence which is $o(1)$.
\\[-2.5mm]
\end{assumption}

\begin{assumption}
 \label{append_assum:parameter}
 The updates of Eq.~\eqref{eq:z_update_n} remain bounded almost surely, i.e.,
\begin{align}
 \sup_n \norm{\zn} < \infty,\quad \textrm{a.s.}
 \notag
\end{align}
\end{assumption}

\begin{lemma}
 \label{lemma:ode_basic_result}
 Assume Assumptions \ref{append_assum:learning_rate}--\ref{append_assum:parameter} hold.
 %
 Let $\zs(\tau),\, \tau\geq s\geq0$, 
denote the trajectory of Eq.~\eqref{eq:z_ode} starting at time $s\in\bbR_{\geq0}$:
 \begin{align}
  \dot\zs(\tau) = \zmap(\zs(\tau)), \quad \forall \tau \in \bbR_{\geq s},
  \notag
 \end{align}
 with $\zs(s) = \bar\z(s)$.
 Then, for any $\T>0$, the following holds almost surely,
 \begin{align}
  &\lim_{s\rightarrow\infty} \sup_{\tau\in[s,s+\T]} \norm{\bar\z(\tau) - \zs(\tau)} = 0,
  \notag
  \\
  &\lim_{s\rightarrow\infty} \sup_{\tau\in[s-\T,s]} \norm{\bar\z(\tau) - \zs(\tau)} = 0.
  \notag
 \end{align}
\end{lemma}

\Proof
This lemma is a simple extension of Lemma 1 in Section 2 of \cite{Borkar08a} with a bias term $\zbias_n$,
and this proof mostly follows from it.
We will only prove the first claim since the same applies to the proof of the second claim.

For $n\in\bbN_{\geq0}$ and $m\in\bbN_{\geq1}$,
%
by the construction, 
$\bar\z$ can be written down as follows:
\begin{align}
 \bar\z(t(n+m)) 
 = 
 \bar\z(t(n)) + \sum_{i=0}^{m-1} \zlr_{n+i}\, \zmap(\bar\z(t(n+i)))
 + \delta_{n,n+m}.
 \label{eq:bar_z}
\end{align}
where 
\begin{align}
 \delta_{n,n+m} &\triangleq \xi_{n+m} - \xi_n + \sum_{i=0}^{m-1} \alpha_{n+i}\, \zbias_{n+i},
 \notag
 \\
 \xi_n &\triangleq 
 \begin{cases}
  0, & \textrm{if\ \ } n=0,
  \\
  \sum_{i=0}^{n-1} \zlr_i \zmds_{i+1}, & \textrm{if\ \ } n\in\bbN_{\geq1}.
 \end{cases}
 \notag
\end{align}

We will show $\sup_{m\geq0}\norm{\delta_{n,n+m}}=0$ as $n\rightarrow\infty$.
By Assumptions \ref{append_assum:mds} and \ref{append_assum:parameter}, the series $\{\xi_n\}$ is a zero mean, 
square-integrable martingale with respect to the $\sigma$-fields $\calF_n$.
Furthermore, by Assumptions \ref{append_assum:learning_rate}, \ref{append_assum:mds}, and \ref{append_assum:parameter}, 
we have 
\[
 \sum_{n=0}^\infty \E[\, \norm{\xi_{n+1} - \xi_n}^2 \given \calF_n ] 
 = \sum_{n=0}^\infty \zlr_n^2 \,\E[\, \norm{\zmds_{n+1}}^2 \given \calF_n] < \infty,\quad  \textrm{a.s.}
\]
From the above and the martingale convergence theorem (Theorem 11 of Appendix in \cite{Borkar08a}), 
it can be said that $\{\xi_n\}$ converges.
The third term of $\delta_{n,n+m}$ also converges to zero as $n\rightarrow\infty$ because $\{\epsilon_n\}$ is $o(1)$ by Assumption \ref{append_assum:bias}.
Thus, the following holds,
\begin{align}
 \lim_{n\rightarrow\infty} \norm{ \delta_{n,n+m}} = 0,\quad  \textrm{a.s.}
 \label{eq:deltaConvergence}
\end{align}

Next, we will look into $\zs$. 
It can be written down as follows:
\begin{align}
 \zs[t(n)](t(n+m)) 
 &= 
 \bar\z(t(n)) + \int_{t(n)}^{t(n+m)} \zmap(\z^{t(n)}(\tau)) d\tau
 \notag
 \\
 &=
 \bar\z(t(n)) 
 + \sum_{i=0}^{m-1} \zlr_{n+i}\, \zmap(x^{t(n)}(t(n+i)))
 + \int_{\tau=t(n)}^{t(n+m)}\! \big( \zmap(\z^{t(n)}(\tau)) - \zmap(\zs[t(n)](\tilde\tau)) \big) d\tau,
 \label{eq:zs}
\end{align}
where
\begin{align}
 \tilde\tau \triangleq \max\{t(n) \given  t(n)\leq\tau,\, n\in\bbN_{\geq0}\}.
 \notag
\end{align}
We investigate the integral on the right-hand side in Eq.~\eqref{eq:zs}.
%
Let $C_0 \triangleq \sup_n \norm{\zn}$.
Note that $C_0 < \infty$ a.s. by Assumption \ref{append_assum:parameter}.
By Assumption \ref{append_assum:lipschitz}, 
$\norm{\zmap(\z) - \zmap(0)} \leq L \norm{\z}$,
and so
\begin{align}
 \norm{\zmap(\z)} \leq \norm{\zmap(0)} + L \norm{\z}.
 \label{eq:zmap_bound}
\end{align}
Therefore, the following holds, for $\tau \in [s, s+\T]$,
\begin{align}
 \norm{\zs(\tau)} 
 &\leq 
 \norm{\bar\z(s)} + \int_{x=s}^\tau \big( \norm{\zmap(0)} + L \norm{\zs(x)}  \big) dx
 \notag
 \\
 &\leq
 C_0 + \norm{\zmap(0)}\T + L \int_{x=s}^\tau \norm{\zs(x)}dx.
 \notag
\end{align}
By Gronwall's inequality (Lemma 6 of Appendix in \cite{Borkar08a}),  we obtain
\begin{align}
 \norm{\zs(\tau)} \leq \big( C_0 + \norm{\zmap(0)}\T)\exp(L\T),\quad 
 \forall \tau \in [s, s+\T].
\end{align}
Thus, from Eq.~\eqref{eq:zmap_bound}, 
we have the following bound,
\begin{align}
 C_\T \triangleq \norm{\zmap(0)} + L ( C_0  + \norm{\zmap(0)}\T)\exp(L\T) < \infty, \quad  \textrm{a.s.}
 \notag
\end{align} 
such that, for all $\tau\in[s,s+\T]$,
\begin{align}
 \norm{\zmap(\zs(\tau))} \leq  C_\T.
 \label{eq:norm_zmap_bound}
\end{align}
Here we assume $\T$ is larger than $t(n+m) - t(n)$ without loss of generality.
For $i \in \{0, \dots, m-1\}$ and $\tau\in[t(n+i), t(n+i+1)]$,
the bound $C_t$ gives
\begin{align}
 \norm{\zs[t(n)](\tau) - \zs[t(n)](t(n+i))} 
 &\leq 
 \normlr{\int_{\iota=t(n+i)}^\tau \zmap(\zs[t(n)](\iota))d\iota}
 \notag
 \\
 &\leq
 C_\T (\tau - t(n+i))
 \notag
 \\
 &\leq
  C_\T \zlr_{n+i}.
 \notag
\end{align}
The inequality gives the bound of the integral in Eq.~\eqref{eq:zs} as follows:
because
\begin{align*}
 \normlr{\int_{\tau=t(n)}^{t(n+m)}\!\!\big(\zmap(\zs[t(n)](\tau)) - \zmap(\zs[t(n)](\tilde\tau))  \big)d\tau}
 &\leq 
 \int_{\tau=t(n)}^{t(n+m)}\!\!L\norm{ \zs[t(n)](\tau) -\zs[t(n)](\tilde\tau) }
 \\
 &=
 L \sum_{i=0}^{m-1} \int_{\tau=t(n+i)}^{t(n+i+1)} \norm{ \zs[t(n)](\tau) -\zs[t(n)](t(n+i)) } d\tau
 \\
 &\leq 
 C_\T L \sum_{i=0}^{m-1} \zlr_{n+i}^2
\end{align*}
Thus, by Assumption \ref{append_assum:learning_rate},
we have
\begin{align}
 \lim_{n\rightarrow\infty} 
 \normlr{\int_{\tau=t(n)}^{t(n+m)}\!\!\big(\zmap(\zs[t(n)](\tau)) - \zmap(\zs[t(n)](\tilde\tau))  \big)d\tau}
 \leq
 C_\T L  \sum_{i=0}^{m-1} \lim_{n\rightarrow\infty} \zlr_{n+i}^2
 =0,
 \quad \textrm{a.s.}
 \label{eq:intConvergence}
\end{align}
%
%
By subtracting Eq.~\eqref{eq:bar_z} from Eq.~\eqref{eq:zs} and taking a norm, 
we have
\begin{align}
 \norm{\bar\z(t(n+m)) - \zs[t(n)](t(n+m))}
 \leq 
 L \sum_{i=0}^{m-1} \zlr_{n+i} \norm{\bar\z(t(n+i)) - \zs[t(n)](t(n+i))} + K_{n,\T},
 \notag
\end{align}
where 
\begin{align*}
 K_{n,\T} &\triangleq C_\T L \sum_{i\geq0}^{\acute m_{n,\!\T}-1} \zlr_{n+i}^2 + \norm{\delta_{n,n+\acute m_{n,\!\T}}},
 \\
 \acute m_{n,\T} &\triangleq  \max\{m \given  t(n+m)\!-\!t(n)\!\leq\!\T,\, m\in\bbN_{\geq0}\}.
\end{align*}
%
%
%
Note that
\begin{align}
 \lim_{n\rightarrow\infty} K_{n,\T} = 0,  \quad \textrm{a.s.}
 \label{eq:K_convergence}
\end{align}
holds by Eqs.~\!\eqref{eq:deltaConvergence} and \eqref{eq:intConvergence}.
By applying the discrete Gronwall lemma (Lemma 8 of Appendix in \cite{Borkar08a}) to the above inequality,
we have
\begin{align}
 \sup_{i\in\{0,\dots,m\}}  \norm{\bar\z(t(n+i)) - \zs[t(n)](t(n+i))}
 \leq 
 K_{n,\T} \exp(L\T),
 \quad \textrm{a.s.}
 \label{eq:norm_z_bound}
\end{align}
Let $\tau\in [t(n+i),t(n+i+1)]$ for $0\leq i \leq m-1$.
Then we have
\[
 \bar\z(\tau) = \lambda \z(t(n+i)) + (1-\lambda)\bar\z(t(n+i+1))
\]
for some $\lambda \in [0,1]$,
and thus the following inequality is obtained,
\begin{align*}
 \norm{\bar\z(\tau) - \zs[t(n)](\tau)}
 &= 
 \norm{\lambda(\bar\z(t(n+i)) - \zs[t(n)](\tau)) + (1+\lambda)(\bar\z(t(n+i+1)) - \zs[t(n)](\tau)  )}
 \\
 &\leq
 \lambda\normlr{\bar\z(t(n+i))  - \zs[t(n)](t(n+i)) - \int_{\iota=t(n+i)}^{\tau}\zmap(\zs[t(n)]) d\iota}
 \\
 &\qquad 
 + (1-\lambda) \normlr{ \bar\z(t(n+i+1)) - \zs[t(n)](t(n+i+1)) + \int_{\iota=\tau}^{t(n+i+1)}\zmap(\zs[t(n)]) d\iota}
 \\
 &\leq
 \lambda\norm{\bar\z(t(n+i))  - \zs[t(n)](t(n+i))} + (1-\lambda) \norm{ \bar\z(t(n+i+1)) - \zs[t(n)](t(n+i+1)) }
 \\
 &\qquad
 + \int_{\iota=t(n+i)}^{t(n+i+1)}\norm{\zmap(\zs[t(n)](\iota))}d\iota
 \\
 &\leq
 K_{n,\T} \exp(L\T) + C_\T \alpha_{n+i},
 \quad \textrm{a.s.},
\end{align*}
where the last inequality is derived by using Eqs.~\!\eqref{eq:norm_z_bound} and \eqref{eq:norm_zmap_bound}.
The above inequality is easily generalized to,
with some constant $C \geq 0$
\[
  \sup_{\tau\in[s,s+\T]} \norm{\bar\z(\tau) - \zs[t(n)](\tau)}
  \leq 
  C K_{\tilde s,\T} \exp(L\T) + C_\T \alpha_{\tilde s},
\]
where $\tilde s \triangleq \max\{t(n) \given  t(n)\leq s,\, n\in\bbN_{\geq0}\}$.
As $s\rightarrow\infty$, we have the first claim in this lemma.
\qed
\\

By applying Lemma \ref{lemma:ode_basic_result} to Theorem 2 of Section 2 and Theorem 2 of Section 6 in \cite{Borkar08a},
we instantly obtain the following lemmas.
\begin{lemma}
 \label{lemma:sa_convergence}
 Assume Assumptions \ref{append_assum:learning_rate}--\ref{append_assum:parameter} hold. 
 Then, the sequence $\{\zn\}$ generated by Eq.~\eqref{eq:z_update_n} almost surely converges to a (possibly sample path dependent) compact connected internally chain transitive invariant set of Eq.~\eqref{eq:z_update_t}.
\end{lemma}
\vspace{3mm}

\begin{lemma}
 \label{lemma:twotimescale_sa_convergence}
 Let the sequence $\{(\xn, \yn)\}$ is generated by
 \begin{align}
 \xn[n+1] &= \xn + \xlr_n [\, \xmap(\xn,\yn) + \xmds_{n+1} + \xbias_n],
  \tag{\ref{eq:x_update_n}}
 \\
 \yn[n+1] &= \yn + \ylr_n [\, \ymap(\xn,\yn) + \ymds_{n+1} + \ybias_n],
  \tag{\ref{eq:y_update_n}}
 \end{align}
 where $\xmap:\bbR^{d}\times\bbR^e\!\rightarrow\bbR^d$ and 
 $\ymap:\bbR^{d}\times\bbR^e\!\rightarrow\bbR^e$ are the expected update functions,
 \mbox{$\xmds\!\in\bbR^d$} and \mbox{$\ymds\!\in\bbR^e$} are noise terms,
 and \mbox{$\xbias\!\in\bbR^d$} and  \mbox{$\ybias\!\in\bbR^e$} are bias terms.
 Also, let the learning rates $\xlr_n$ and $\ylr_n$ of Eqs.~\!\eqref{eq:x_update_n} and \eqref{eq:y_update_n} satisfying
\begin{align}
 \left\{
 \begin{array}{l}
  \displaystyle \
  \lim_{N\rightarrow\infty} \sum_{n=0}^N \xlr_n
  = \lim_{N\rightarrow\infty} \sum_{n=0}^N \ylr_n
  = \infty,
  \\[5mm]
  \displaystyle \
  \lim_{N\rightarrow\infty} \sum_{n=0}^N \big( \xlr_n^2 + \ylr_n^2 \big) 
  < \infty ,
  \\[5mm]
  \displaystyle \
  \lim_{N\rightarrow\infty}\frac{\xlr_N}{\ylr_N} = 0.
 \end{array}
\right.
 \notag
\end{align}
 Assume Assumptions  Assumptions 1--5 hold. 
 Then, the sequence $\{(\xn, \yn)\}$  almost surely converges to a (possibly sample path dependent) compact connected internally chain transitive invariant set $A$ of the following ODEs
\begin{align}
 \dot\x(\tau) &=  \xmap(\x(\tau),\y(\tau)),
 \tag{\ref{eq:x_update_t}}
 \\
 \dot\y(\tau) &=  \ymap(\x(\tau),\y(\tau)).
 \tag{\ref{eq:y_update_t}}
\end{align}
 Any pair $(\x, \y) \in A$ has the relation $\y = \varphi(\x)$, where $\varphi(\x)$ is defined in Assumption 4
 and denotes the globally asymptotically stable equilibrium of the ODE \eqref{eq:y_update_t} of $\y$ given $\x$.
\end{lemma}

\subsection{Propositions 1 and 2}

By applying Lemma \ref{lemma:twotimescale_sa_convergence} 
to the update rule of the proposed PG-MCTL algorithm (Eqs.~\!\eqref{eq:x_update_n} and \eqref{eq:y_update_n}),
we immediately obtain Propositions \ref{prop:mfun_invariant} and \ref{prop:mfun_adaptive}.
\\[-1mm]

\newtheorem*{propositionOne}{Proposition 1}
\begin{propositionOne}
 Assume Assumptions 1--5 hold.
 Let the mixing probability function $\mfunn:\calH\rightarrow[0,1]$ be invariant to the number of episodes $n$
 and the learning rates $\xlr_n$ and $\ylr_n$ satisfying
\begin{align}
 \left\{
 \begin{array}{l}
  \displaystyle \
  \lim_{N\rightarrow\infty} \sum_{n=0}^N \xlr_n
  = \lim_{N\rightarrow\infty} \sum_{n=0}^N \ylr_n
  = \infty,
  \\[5mm]
  \displaystyle \
  \lim_{N\rightarrow\infty} \sum_{n=0}^N \big( \xlr_n^2 + \ylr_n^2 \big) 
  < \infty ,
  \\[5mm]
  \displaystyle \
  \lim_{N\rightarrow\infty}\frac{\xlr_N}{\ylr_N} = 0.
 \end{array}
\right.
 \tag{13} 
\end{align}
Then, almost surely, the sequence $\{(\xn,\yn)\}$ generated by Eqs.~\eqref{eq:x_update_n} and \eqref{eq:y_update_n}
converges to a  compact connected internally chain transitive invariant set of 
Eqs.~\eqref{eq:x_update_t} and \eqref{eq:y_update_t}.
\end{propositionOne}
\vspace{3mm}

\newtheorem*{propositionTwo}{Proposition 2}
\begin{propositionTwo}
 Let $\mfunn:\calH\rightarrow[0,1]$ be a function parameterized by a part of $\x$
 (and be a Lipschitz continuous map with respect to its parameter).
 Assume that all the conditions of Proposition 1 are satisfied 
 expect for $\mfunn$.
 Still, the consequence of Proposition 1 holds.
\end{propositionTwo}

\subsection{Proposition 3}

\newtheorem*{propositionThree}{Proposition 3}
\begin{propositionThree}
 Assume  Assumptions \ref{assum:noise} and \ref{assum:bias} only for $i=1$ hold,
 $\xmap$ is Lipschitz continuous map, and $\sup_n(\norm{\xn})<\infty$ holds.
 Let the mixing probability $\mfunn$ be $o(1)$ 
 and satisfy $0\leq \mfunn \leq 1-\varepsilon$ for all $n$ and a constant $\varepsilon>0$, 
 and the learning rate of the PG policy satisfy
\begin{align}
  \lim_{N\rightarrow\infty} \sum_{n=0}^N \xlr_n  = \infty, \quad
  \lim_{N\rightarrow\infty} \sum_{n=0}^N  \xlr_n^2   < \infty.
\notag
\end{align}
Then, 
almost surely, the sequence $\{\xn\}$ generated by Eqs.~\!\eqref{eq:x_update_n} and \eqref{eq:y_update_n}
converges to a compact connected internally chain transitive invariant set of the ODE, $\dot\x(\tau) = \difx \obj(\polxt)$.
\end{propositionThree}

\Proof
The update rule of $\x$ (Eq.~\!\eqref{eq:x_update_n} ) is rewritten as
\begin{align}
 \xn[n+1] &= \xn+ \xlr_n \sum_{t=0}^\tmax \difx\log \polxyn(a_t|h_t) (\ret_t - b(h_t))
 \notag
 \\
 &=
 \xn + \xlr_n 
 \left(
 \sum_{t=0}^\tmax \difx\log \polxn(a_t|h_t) (\ret_t - b(h_t))
 -  \lambda_t \sum_{t=0}^\tmax \polyn(a_t|h_t)\difx\log \polxn(a_t|h_t) (\ret_t - b(h_t))
 \right)
 \label{eq:x_update_n_2}
\end{align}
By the definition of the PG-MCTL policy (Eq.~\!\eqref{eq:polxy}) 
\begin{align}
 \polxyn(a|h) &\triangleq (1-\mfun_n) \polxn(a|h) + \mfun_n \polyn(a|h)
 \notag
\end{align}
and the assumption of the proposition, $1-\mfun_n\geq \varepsilon,\ n\geq0$,
the expected value of the second terms of the right side of Eq.~\!\eqref{eq:x_update_n_2}
is
\begin{align}
 \Epolxyn\!\!
 &\left[\
 \sum_{t=0}^\tmax \difx\log \polxn(A_t|H_t) (\Ret_t - b(H_t))
 -  \lambda_t \sum_{t=0}^\tmax \polyn(A_t|H_t)\difx\log \polxn(A_t|H_t) (\Ret_t - b(H_t))
 \right]
 \notag
 \\
 &= 
 \varepsilon^\tmax 
 \Epolxn
 \!\!
 \left[\, \sum_{t=0}^\tmax\difx \log \polx (A_t, H_t)\, (\Ret_t - b(h_t))  \right]
 +
 \epsilon'_t,
 \notag
\end{align}
where the sequence $\{\epsilon'_n\}$ is $o(1)$.
Thus, Eq.~\!\eqref{eq:x_update_n_2} 
 can be rewritten as
\begin{align}
 \xn[n+1] &= \xn + \alpha_n (\xmapp(\xn) + M'_n + \epsilon'_n),
 \notag
\end{align}
where $\xmapp:\bbR^d\rightarrow\bbR^d$ is the expected update function
\begin{align}
 \xmapp(\x) 
 \triangleq 
 \varepsilon^\tmax\!\! \left[\, \sum_{t=0}^\tmax\difx \log \polx (A_t, H_t)\, (\Ret_t - b(h_t))  \right]\!
 = 
 \varepsilon^\tmax
 \difx \obj(\polxt),
 \notag
\end{align}
and $\{M'_n\}$ is a zero mean, square-integrable martingale difference sequence with respect to $\calF_n$.

From the above, we can apply Lemma \ref{lemma:sa_convergence} and so the claim follows.
\qed

\subsection{Theorem 1}

\newtheorem*{theoremOne}{Theorem 1}
\begin{theoremOne}
 Let the PG-MCTL update the parameterized policy $\polx$ by Eq.~\!\eqref{eq:pg_update_mixpolicy}
 and the MCTL policy $\poly$ by the rule in which  $\kappa$ in Eq.~\!\eqref{eq:mcts_update_2} 
 is replaced by $\bar\kappa$ of Eq.~\!\eqref{eq:bounded_adjustment},
 and the learning rates satisfy the conditions of Eq.~\!\eqref{eq:two_timescale_learning_rate}.
 Also let $\polx$ be defined on a compact parameter space and have always bounded first and second partial derivatives, 
 and $\poly$ be a softmax policy with hyper parameters $\beta\geq 0$ and $C\geq0$ as
 \begin{align}
  \poly(a|h) \propto
  \exp\!\!
  \left(\!
  \beta \Bigg\{
  \qhat(h,a)+C\!\sqrt{ \invvst(h,a) \log\!\bigg(\!\sum_b \frac{1}{\invvst(h,b)} \bigg) }
  \Bigg\}
  \!\right).
  \tag{\ref{eq:softUCT}} 
 \end{align}
 Then, 
 $\lim_{n\rightarrow\infty} \difx \obj(\polxyn)=0$ holds.
\end{theoremOne}
\Proof
The proof consists of two major steps.
First, we will show that the parameter $\{(\xn,\yn)\}$ converges to 
a compact connected internally chain transitive invariant set.
Then we will prove that any element in that set satisfies the properties claimed in the theorem.

To apply Lemma \ref{lemma:twotimescale_sa_convergence},
we will investigate whether the conditions of Lemma \ref{lemma:twotimescale_sa_convergence} are satisfied.
By the construction of the sequence  $\{\yn\}$ of the parameter of the MCTL policy,
\begin{align}
 \sup_{n} \norm{\yn} < \infty  
 \notag
\end{align}
holds.
It means that Assumption 5 holds, taking into account the condition $\sup_n\norm{\xn}<\infty$,
and also ensures that $\poly$ always has bounded first and second derivatives, as well as $\polx$.
In order to check Lipschitz continuity of the expected update functions $\xmap(\x,\y)$ and $\ymap(\x,\y)$ 
(Assumption \ref{assum:update_function}),
we define them as
\begin{align}
 \xmap(\x,\y)
 &\triangleq 
 \Epolxy\!\left[
 \, \sum_{t=0}^\tmax \difx \log \polxy(A_t|H_t) (\Ret_t - b(H_t))
 \right]
 =\difx \obj(\polxy),
 \label{eq:xmap_pgmcts}
\end{align}
and
\begin{align}
\begin{cases}
 \displaystyle
 [\ymap(\x,\y)]_{\invvst(h_t,a)} 
 \triangleq
 - \visitprob(h_t,a)  \frac{\bar\kappa_{\ychar}(h_t,a) \invvst(h_t,a)}{\invvst(h_t,a)+1},
 \ \forall (h_t,a) \in \calH_t\times\calA,\ \  \forall t\in\{0,\dots,\tmax\},
 \\
 \displaystyle
 [\ymap(\x,\y)]_{\qhat(h_t,a)}
 \triangleq
 \visitprob(h_t,a) \bar\kappa_{\ychar}(h_t,a)
 \big( \Epolxy[\Ret_t \given H_t\!=\!h_t,A_t\!=\!a] - \qhat(h_t,a)
 \big),
 \ifpreprint 
 \ \ \forall(h_t,a) \in \calH_t\times\calA,\ \  \forall t\in\{0,\dots,\tmax\},
 \else 
 \\ \hspace{7.25cm} 
 \forall(h_t,a) \in \calH_t\times\calA,\ \  \forall t\in\{0,\dots,\tmax\},
 \fi 
\end{cases}
 \label{eq:ymap_pgmcts}
\end{align}
where $[\ymap(\x,\y)]_{x}$ denotes the output corresponding to the parameter $x$ in $\y$,
the function $\visitprob(h_t,a)$ is the experiencing probability of $(h_t,a)$ under $\polxy$,
\begin{align}
 \visitprob(h_t,a) &\triangleq \Pr(H_t\!=\!h_t, A_t\!=\!a \given \Env, \polxy),
 \notag
\end{align}
and $\bar\kappa_{\ychar}$ is the counterpart to $\bar\kappa_{n,t}$ defined in Eq.~\eqref{eq:bounded_adjustment}.
Thus, with the above properties and the boundedness of the reward function, we can see that
$\xmap$ and $\ymap$ are Lipschitz continuous maps, i.e., Assumption 3 holds.
The above observations also indicate that Assumptions 1 and 2 hold.
Furthermore, since the second term in the MCTL policy (Eq.~\eqref{eq:softUCT}) is asymptotically negligible,
our task is episodic, and $\x$ is basically updated with a naive Monte Carlo method, 
the ODE corresponding to $\ymap$ has a globally asymptotically stable equilibrium $\varphi(\x)$, 
which will depend on $\x$,
i.e., Assumption 4 holds.
From the above results,
Lemma \ref{lemma:twotimescale_sa_convergence} can be applied to this setup.
Thus, it is proven that $\{(\xn,\yn)\}$ converges to 
a compact connected internally chain transitive invariant set $\calS$ of the ODEs
corresponding to Eqs.~\!\eqref{eq:xmap_pgmcts} and \eqref{eq:ymap_pgmcts}, 
and $\y = \varphi(\x)$ holds for all $(\x,\y)\in \calS$.

With the above observations,
we can instantly prove 
$\lim_{n\rightarrow\infty} \difx \obj(\polxyn)=0$ by contradiction.
(This is because $\xn$ converges to
 a compact connected internally chain transitive invariant set 
of the ODE $\difx \obj(\polxy)$ and 
$\obj(\polxy)$ is bounded by the HDP definition.)

%
%
%
%
%
%
\qed

 \subsection{Equivalence of the standard MCTS update and Eq.~(\ref{eq:mcts_update_2})}
\label{appendix:equivalence}
By construction of $\invvst(h,a)$ in Eq.~\eqref{eq:mcts_update_2}, the initial value $\invvst$ is $1$ for all $(h,a)$.
If $\invvst$ is updated once or more than once at $(h,a)$, $\invvst(h,a)$ is equal to or less than $0.5$.
Thus, by the definition of the tree-inclusion probability $p_{n,t}$ in Eq.~\eqref{eq:treeIncProb},
if $\invvst(h_{t-1},a_{t-1})$ has been updated even once in past episodes,
the tree-inclusion probability $p_{n,t}(h_t,a_t)$ is one,
otherwise it is zero.
%
It indicates that in addition to the case $t=0$, as long as 
a node corresponding to $(h_{t-1},a_{t-1}),\, t\in\bbN_{\geq1},$ would be included in a tree if the standard MCTS update were used, 
the tree-inclusion probability $p_{n,t}(h_t,a_t)$ is $1$,
and thus $\invvst$ and $\qhat$ of $(h_t,a_t)$ will be updated with probability 1.
Otherwise, Eq.~\eqref{eq:mcts_update_2} will not change $\invvst$ and $\qhat$ of $(h_t,a_t)$ at all.
The above is the same as the standard MCTS update.

%
%
%

%
All that remains is to show that the update rule in Eq.~\eqref{eq:mcts_update_2} 
can be derived from the standard MCTS update rule in Eq.~\!\eqref{eq:mcts_update} when $p_{n,t}=1$.
%
Because of $\ylr_{n} \triangleq 1/n$ and 
$\kappa_{n,t} \triangleq  p_{n,t} \invvst_n(h_t,a_t)/\ylr_n =  n\invvst_n(h_t,a_t) $,
the update of $m$ in Eq.~\!\eqref{eq:mcts_update} can be transformed as
\newcommand{\mnn} {m_{n+1}(h_t,a_t)}
\newcommand{\mn}  {m_{n}(h_t,a_t)}
\newcommand{\unn} {\invvst_{n+1}(h_t,a_t)}
\newcommand{\un}  {\invvst_{n}(h_t,a_t)}
\newcommand{\qnn} {q_{n+1}(h_t,a_t)}
\newcommand{\qn} {q_{n}(h_t,a_t)}
\begin{alignat}{2}
 & \mnn
 &\ =\ &
 \mn + 1
 \notag
 \\ 
 \Leftrightarrow \hspace{5mm} &
 \frac{1}{\unn} 
 &\ =\ & \frac{1}{\un} + 1
 \notag
 \\
 \Leftrightarrow \hspace{5mm} &
 \unn
 &\ =\ & \frac{\un}{1+\un}
 \notag
 \\
 &&\ =\ & \frac{\un(1+\un)-\un^2}{1+\un}
 \notag
 \\
 &&\ =\ & \un -\frac{\un^2}{1+\un}
 \notag
 \\
 &&\ =\ & \un - \frac{1}{n} n \un \frac{\un}{1+\un}
 \notag
 \\
 &&\ =\ & \un +  \ylr_n \kappa_{n,t} \frac{-\un}{1+\un}.
 \notag
\end{alignat}
The update of $q$ in Eq.~\!\eqref{eq:mcts_update} can also be transformed into 
\begin{align}
 \qnn
 &=
 \qn + \frac{1}{m_n(h_t,a_t)} ( \ret_t - \qhat_n(h_t,a_t)),
 \notag
 \\ 
 &=
 \qn +  \un ( \ret_t - \qhat_n(h_t,a_t)),
 \notag
 \\ 
 &=
 \qn + \frac{1}{n} n \un  ( \ret_t - \qhat_n(h_t,a_t)),
 \notag
 \\
 &=
 \qn +  \ylr_n \kappa_{n,t} ( \ret_t - \qhat_n(h_t,a_t)).
 \notag
\end{align}
Eq.~\eqref{eq:mcts_update_2} is derived.
\qed

\section{Experimental setup}
\label{appendix:exp_setup}

\subsection{Randomly synthesized task}
\label{appendix:random_task}
The first test problem is a non-Markovian task that is a simple but illustrative HDP.
It is randomly synthesized to be analogous to a generation task such as text generation and compound synthesis.
There are $5$ observations and $10$ actions,
i.e., $\abs{\calO}=5$ and $\abs{\calA}=10$.
The observation probability function $\po$, 
which corresponds to the history transition probability $\ph$,
was synthesized to depend on the time-step, observation, and action.
Specifically, a probability vector for the observation was generated by 
the Dirichlet distribution $\textrm{Dir}(\bm\alpha=[0.2,\dots,0.2])$
independently for each $(t, o,a)$.
The reward function $\rfun$ was synthesized to 
have the following structure:
\[
 \rfun(h_t,a_t) = 
 \begin{cases}
  \frac{1}{\tmax} x(o_t, a_t)
  &\! \textrm{if\ } t < \tmax,
  \\
  y(h_t) + 10 z(o_0,o_1,\dots,o_t) 
  &\! \textrm{otherwise},
 \end{cases}
\]
where $x$ and $y$ are the per-step reward function and 
the history-based reward function, respectively.
Each value of those functions was initialized independently by the normal distribution
$\mathcal{N}(\mu\!=\!0,\sigma^2\!=\!1)$.
The values of the function $z$ were set by using a Gaussian process
so that the more similar the observation series $(o_0, \dots,o_{\tmax})$ 
and $(o'_0,\dots,o'_{\tmax})$ were, the closer 
$z(o_0, \dots,o_{\tmax})$ and $z(o'_0, \dots,o'_{\tmax})$ tended to be.
Its covariance function was defined with Hamming distance,
and the variance was equal to $1$.

This reward function $\rfun$ can be interpreted in the context of text generation as follows.
The function $z$, which is a dominant part in $\rfun$, represents the quality of the generated text, 
$x$ represents the quality of local word connections,
and $y$ is like noise.

The policy $\polx$ was a softmax and parameterized to have the same structure as the reward function.
It will correspond to using domain knowledge and 
will be a usual setting since the reward function is often predefined by the user. 
Specifically, $\polx$ had a parameter for each $(o_t,a_t)$, $(o_0,o_1,..,o_t,a_t)$, and $(h_t,a_t))$,
though it was a redundant parameterization.
The hyper-parameters of the applied algorithms are shown in Table~\ref{tab:hpara_random_task}.
As described in Section \ref{sec:experiments}, for fair evaluation,
we ﬁrst tuned the hyper-parameters of the REINFORCE and MCTL algorithms 
and then used them for the PG-MCTL and the naive mixture algorithms.
The hyper-parameters of the lazy AlphaZero were tuned independently.

It should be noted that the experiments here were conducted on an ordinary Laptop, 
and the computation time was only a few days.

\begin{table}[ht]
 \caption{Hyper-parameters used in the randomly synthesized task.}
  \label{tab:hpara_random_task}
  \centering
  \begin{tabular}{lccccc}
    \toprule
   \textbf{Algorithm} & $\alpha$ & $C$ & $\lambda$ & $\beta$ & $M$\\
    \midrule
    REINFORCE      & $0.01$  &   -  &   -    &   -   &   -    \\
    Naive mixture  & $0.01$  & $5$  & $0.2$  &   -   & - \\
    MCTL           & -       & $5$  &   -    &   -   &   -    \\
    Lazy AlphaZero & $0.0067$ & $15$ &   -   &   -   &   -    \\
    PG-MCTL        & $0.01$  & $5$  & $0.2$  & $100$ & $50000$\\
    PG-MCTL-adpt   & $0.01$  & $5$  &   & $100$ & $50000$\\
    \bottomrule
  \end{tabular}
\end{table}

\subsection{T-maze task}
%
%
%
%


%
There are four possible actions: moving north, east, south, or west. 
At the initial time step $t=0$, the agent starts at position $\textsf S$
and perceives a signal $\textsf{X}\in\{1000, 0100\}$, indicating whether the goal position $\textsf G$
is situated on either the north or south side of the T-junction.
At each subsequent time step $t \in \{1, 2, \dots\}$, the agent observes its current location type.
In the corridor, the observation is $0010$.
At the T-junction, the observation is $0001$, lacking any information about the goal's position.
Therefore, the agent must memorize the initial observation to navigate effectively.

The reward settings are as follows.
If the correct action is chosen at the T-junction,
e.g., move north if $\textsf{X}$ is $1000$ and south if $0100$,
the agent receives a reward of $4.0$, otherwise a reward of $-0.1$. 
In both cases, the episode ends. 
When it is in the corridor and chooses to move north or south,
it stays there and receives a reward of $-0.1$.
Otherwise, the reward will be zero.

The settings for our model are as follows: 
The LSTM network, following standard architectures used in reinforcement learning \cite{Bakker02a, Wierstra10a}, 
takes four input units corresponding to the observation dimensions and processes them through a hidden layer with eight memory cells. 
The LSTM's output is concatenated with the original observation, to form a feature vector.
This feature vector is subsequently utilized in linear layers to compute the action values $q$ and the baseline $b$. 
This design aims to capture temporal dependencies and learn effective representations for HDPs.

The hyper-parameters of the applied algorithms are listed in Table~\ref{tab:hpara_tmaze_task}.
The discount rate for cumulative rewards was set to $\gamma=0.98$.
The means of selecting the hyper-parameters is the same as 
for the randomly synthesized task (see \ref{appendix:random_task}).
However, while it was necessary to reduce the hyper-parameter $C$ to $0.1$ for MCTL to obtain the optimal policy, 
this setting required a large number of episodes.
We therefore set $C$ to a slightly larger value of $0.3$ to balance the learning accuracy and speed.
On the other hand, for both PG-MCTL and PG-MCTL-adpt, the hyper-parameter $C$ remained at $0.1$, 
as these methods did not necessitate an excessive number of episodes with this configuration.

It is noteworthy that these experiments were conducted on a public cloud,
utilizing a single NVIDIA Tesla T4 GPU, 
with the total computation time being approximately one week.


{
\begin{table}[t]
 \caption{Hyper-parameters used in the T-maze task.}
  \label{tab:hpara_tmaze_task}
  \hspace{-10mm}
  \begin{tabular}{lcccccccccccccc}
    \toprule
   & \multicolumn{7}{c}{$L=30$, $s_0=0$}  
       & \multicolumn{7}{c}{$L=100$, $s_0=50$} \\
   \cmidrule(r){2-8}   \cmidrule(r){9-15}
   \textbf{Algorithm} 
   & $\alpha$ & $C$ & $\lambda$ & $\beta$ & $M$ & epoch & clip
   & $\alpha$ & $C$ & $\lambda$ & $\beta$ & $M$ & epoch & clip \\
   \midrule
    REINFORCE        & $0.2$  &   -   &   -   &   -   &   -    &   -   &   -    
                     & $0.1$  &   -   &   -   &   -   &   -    &   -   &   -    \\
    PPO              & $0.06$ &   -   &   -   &   -   &   -   &   3   &   0.2    
                     & $0.03$ &   -   &   -   &   -   &   -   &   3   &   0.2    \\
    MCTL              &  -     & $0.3$ &   -   &   -   &   -    &   -   &   -    
                     &  -     & $0.3$ &   -   &   -   &   -    &   -   &   -    \\
    Lazy AlphaZero   & $0.02$ & $1$   &   -   &   -   &   -    &   -   &   -    
                     & $0.01$ & $1$   &   -   &   -   &   -    &   -   &   -    \\
    PG-MCTL          & $0.2$  & $0.1$ & $0.2$ & $100$ & $3000$ &   -   &   -    
                     & $0.1$  & $0.1$ & $0.2$ & $100$ & $5000$ &   -   &   -    \\
    PG-MCTL-adpt     & $0.2$  & $0.1$ & -     & $100$ & $3000$ &   -   &   -    
                     & $0.1$  & $0.1$ & -     & $100$ & $5000$ &   -   &   -    \\
    \bottomrule
  \end{tabular}
\end{table}
}

%% file: main.bbl
\begin{thebibliography}{88}
\providecommand{\natexlab}[1]{#1}
\providecommand{\url}[1]{\texttt{#1}}
\expandafter\ifx\csname urlstyle\endcsname\relax
  \providecommand{\doi}[1]{doi: #1}\else
  \providecommand{\doi}{doi: \begingroup \urlstyle{rm}\Url}\fi

\bibitem[Aberdeen(2003)]{AberdeenThesis}
D.~Aberdeen.
\newblock \emph{Policy-Gradient Algorithms for Partially Observable {M}arkov
  Decision Processes}.
\newblock PhD thesis, Australian National University, 2003.

\bibitem[Anthony et~al.(2017)Anthony, Tian, and Barber]{Anthony17a}
T.~Anthony, Z.~Tian, and D.~Barber.
\newblock Thinking fast and slow with deep learning and tree search.
\newblock In \emph{Advances in Neural Information Processing Systems}, 2017.

\bibitem[Anthony et~al.(2019)Anthony, Nishihara, Moritz, Salimans, and
  Schulman]{Anthony19a}
T.~Anthony, R.~Nishihara, P.~Moritz, T.~Salimans, and J.~Schulman.
\newblock Policy gradient search: Online planning and expert iteration without
  search trees.
\newblock In \emph{arXiv preprint arXiv:1904.03646}, 2019.

\bibitem[Bacchus et~al.(1996)Bacchus, Boutilier, and Grove]{Bacchus96a}
F.~Bacchus, C.~Boutilier, and A.~Grove.
\newblock Rewarding behaviors.
\newblock In \emph{National Conference on Artificial Intelligence}, volume~2,
  pp.\  1160–1167. AAAI Press, 1996.

\bibitem[Bacchus et~al.(1997)Bacchus, Boutilier, and Grove]{Bacchus97a}
F.~Bacchus, C.~Boutilier, and A.~Grove.
\newblock Structured solution methods for non-{M}arkovian decision processes.
\newblock In \emph{National Conference on Artificial Intelligence}, pp.\
  112–117. AAAI Press, 1997.

\bibitem[Bakker(2002)]{Bakker02a}
B.~Bakker.
\newblock Reinforcement learning with long short-term memory.
\newblock In \emph{Advances in Neural Information Processing Systems}, 2002.

\bibitem[Baxter \& Bartlett(2001)Baxter and Bartlett]{baxter01a}
J.~Baxter and P.~Bartlett.
\newblock Infinite-horizon policy-gradient estimation.
\newblock \emph{Journal of Artificial Intelligence Research}, 15:\penalty0
  319--350, 2001.

\bibitem[Bellemare et~al.(2016)Bellemare, Srinivasan, Ostrovski, Schaul,
  Saxton, and Munos]{Bellemare16a}
M.~G. Bellemare, S.~Srinivasan, G.~Ostrovski, T.~Schaul, D.~Saxton, and
  R.~Munos.
\newblock Unifying count-based exploration and intrinsic motivation.
\newblock In \emph{Advances in Neural Information Processing Systems}, pp.\
  1471--1479, 2016.

\bibitem[Berg et~al.(2012)Berg, Patil, and Alterovitz]{Berg12a}
J.~Berg, S.~Patil, and R.~Alterovitz.
\newblock Motion planning under uncertainty using iterative local optimization
  in belief space.
\newblock In \emph{International Journal of Robotics Research}, pp.\
  1263–--1278, 2012.

\bibitem[Bertsekas(1995)]{bertsekas95a}
D.~P. Bertsekas.
\newblock \emph{Dynamic Programming and Optimal Control, Volumes 1 and 2}.
\newblock Athena Scientific, 1995.

\bibitem[Bertsekas \& Tsitsiklis(1996)Bertsekas and Tsitsiklis]{bertsekas96a}
D.~P. Bertsekas and J.~N. Tsitsiklis.
\newblock \emph{Neuro-Dynamic Programming}.
\newblock Athena Scientific, 1996.

\bibitem[Borkar(2008)]{Borkar08a}
V.~Borkar.
\newblock \emph{Stochastic Approximation: A Dynamical Systems Viewpoint}.
\newblock Cambridge University Press, 2008.

\bibitem[Brafman \& Giacomo(2019)Brafman and Giacomo]{Brafman19a}
R.~I. Brafman and G.~D. Giacomo.
\newblock Regular decision processes: A model for non-{M}arkovian domains.
\newblock In \emph{International Joint Conference on Artificial Intelligence},
  pp.\  5516--5522, 2019.

\bibitem[Browne et~al.(2012)Browne, Powley, Whitehouse, Lucas, Cowling,
  Rohlfshagen, Tavener, Perez, Samothrakis, and Colton]{Browne12a}
C.~Browne, E.~Powley, D.~Whitehouse, S.~Lucas, P.~I. Cowling, P.~Rohlfshagen,
  S.~Tavener, D.~Perez, S.~Samothrakis, and S.~Colton.
\newblock A survey of {M}onte {C}arlo tree search methods.
\newblock \emph{IEEE Transactions on Computational Intelligence and AI in
  Games}, 4\penalty0 (1):\penalty0 1–--43, 2012.

\bibitem[Burda et~al.(2019)Burda, Edwards, Pathak, Storkey, Darrell, and
  Efros]{Burda19a}
Y.~Burda, H.~Edwards, D.~Pathak, A.~Storkey, T.~Darrell, and A.~A. Efros.
\newblock Large-scale study of curiosity-driven learning.
\newblock In \emph{International Conference on Learning Representations}, 2019.

\bibitem[Chen et~al.(2019)Chen, Beutel, Covington, Jain, Belletti, and
  Chi]{Chen19a}
M.~Chen, A.~Beutel, P.~Covington, S.~Jain, F.~Belletti, and E.~Chi.
\newblock Top-k off-policy correction for a reinforce recommender system.
\newblock In \emph{International Conference on Web Search and Data Mining},
  2019.

\bibitem[Choshen et~al.(2020)Choshen, Fox, Aizenbud, and Abend]{Choshen20a}
L.~Choshen, L.~Fox, Z.~Aizenbud, and O.~Abend.
\newblock On the weaknesses of reinforcement learning for neural machine
  translation.
\newblock In \emph{International Conference on Learning Representations}, 2020.

\bibitem[Ciosek \& Whiteson(2020)Ciosek and Whiteson]{Ciosek20a}
K.~Ciosek and S.~Whiteson.
\newblock Expected policy gradients for reinforcement learning.
\newblock \emph{Journal of Machine Learning Research}, 21:\penalty0 1--51,
  2020.

\bibitem[Clarke et~al.(2015)Clarke, Friedrich, Tartaglia, Marchesotti, Senn,
  and Herzog]{Clarke15a}
A.~M. Clarke, J.~Friedrich, E.~M. Tartaglia, S.~Marchesotti, W.~Senn, and M.~H.
  Herzog.
\newblock Human and machine learning in non-{M}arkovian decision making.
\newblock \emph{PLOS One}, 10\penalty0 (4):\penalty0 e0123105, 2015.

\bibitem[Cou{\"{e}}toux et~al.(2011)Cou{\"{e}}toux, Hoock, Sokolovska, Teytaud,
  and Bonnard]{Couetoux11a}
A.~Cou{\"{e}}toux, J.-B. Hoock, N.~Sokolovska, O.~Teytaud, and N.~Bonnard.
\newblock Continuous upper confidence trees.
\newblock In \emph{International Conference on Learning and Intelligent
  Optimization}, pp.\  433–445, 2011.

\bibitem[Coulom(2006)]{Coulom06a}
R.~Coulom.
\newblock Efficient selectivity and backup operators in {Monte-Carlo} tree
  search.
\newblock In \emph{International Conference on Computers and Games}, pp.\
  72--83, 2006.

\bibitem[Dam et~al.(2021)Dam, D'Eramo, Peters, and Pajarinen]{Dam21a}
T.~Dam, C.~D'Eramo, Jan Peters, and J.~Pajarinen.
\newblock Convex regularization in {Monte-Carlo} tree search.
\newblock In \emph{International Conference on Machine Learning}, pp.\
  2365--2375, 2021.

\bibitem[Dieb et~al.(2020)Dieb, Song, Yin, and Ishii]{Dieb20a}
S.~Dieb, Z.~Song, W.~J. Yin, and M.~Ishii.
\newblock Optimization of depth-graded multilayer structure for {X}-ray optics
  using machine learning.
\newblock \emph{Journal of Applied Physics}, 128\penalty0 (7):\penalty0 074901,
  2020.

\bibitem[Doshi-Velez et~al.(2015)Doshi-Velez, Pfau, Wood, and
  Roy]{DoshiVelez15a}
F.~Doshi-Velez, D.~Pfau, F.~Wood, and N.~Roy.
\newblock Bayesian nonparametric methods for partially-observable reinforcement
  learning.
\newblock \emph{IEEE Transactions on Pattern Analysis and Machine
  Intelligence}, 37\penalty0 (2):\penalty0 394–--407, 2015.

\bibitem[Efroni et~al.(2019)Efroni, Dalal, Scherrer, and Mannor]{Efroni19a}
Y.~Efroni, G.~Dalal, B.~Scherrer, and S.~Mannor.
\newblock How to combine tree-search methods in reinforcement learning.
\newblock In \emph{AAAI Conference on Artificial Intelligence}, pp.\
  3494--3501, 2019.

\bibitem[Friedrich et~al.(2011)Friedrich, Urbanczik, and Senn]{Friedrich11a}
J.~Friedrich, R.~Urbanczik, and W.~Senn.
\newblock Spatio-temporal credit assignment in neuronal population learning.
\newblock \emph{PLOS Computational Biology}, 7\penalty0 (6):\penalty0 e1002092,
  2011.

\bibitem[Grill et~al.(2020)Grill, Altch{\'{e}}, Tang, Hubert, Valko,
  Antonoglou, and Munos]{Grill20a}
J.~B. Grill, F.~Altch{\'{e}}, Y.~Tang, T.~Hubert, M.~Valko, I.~Antonoglou, and
  R.~Munos.
\newblock {Monte-Carlo} tree search as regularized policy optimization.
\newblock In \emph{International Conference on Machine Learning}, pp.\
  3769--3778, 2020.

\bibitem[Grooten et~al.(2022)Grooten, Wemmenhove, Poot, and
  Portegies]{Grooten22a}
B.~Grooten, J.~Wemmenhove, M.~Poot, and J.~Portegies.
\newblock Is vanilla policy gradient overlooked? analyzing deep reinforcement
  learning for hanabi.
\newblock In \emph{Adaptive and Learning Agents Workshop at AAMAS}, 2022.

\bibitem[Gullapalli(1990)]{gullapalli90a}
V.~Gullapalli.
\newblock A stochastic reinforcement learning algorithm for learning
  real-valued functions.
\newblock \emph{Neural Networks}, 3\penalty0 (6):\penalty0 671--692, 1990.

\bibitem[Guo et~al.(2014)Guo, Singh, Lee, Lewis, and Wang]{Guo14a}
X.~Guo, S.~Singh, H.~Lee, R.~L. Lewis, and X.~Wang.
\newblock Deep learning for real-time atari game play using offline
  {Monte-Carlo} tree search planning.
\newblock In \emph{Advances in Neural Information Processing Systems}, 2014.

\bibitem[Guo et~al.(2016)Guo, Singh, Lewis, and Lee]{Guo16a}
X.~Guo, S.~Singh, R.~Lewis, and H.~Lee.
\newblock Deep learning for reward design to improve {Monte Carlo} tree search
  in {ATARI} games.
\newblock In \emph{International Joint Conference on Artificial Intelligence},
  2016.

\bibitem[Haarnoja et~al.(2017)Haarnoja, Tang, Abbeel, and Levine]{Haarnoja17a}
T.~Haarnoja, H.~Tang, P.~Abbeel, and S.~Levine.
\newblock Reinforcement learning with deep energy-based policies.
\newblock In \emph{International Conference on Machine Learning}, volume~70,
  pp.\  1352--1361, 2017.

\bibitem[Haarnoja et~al.(2018)Haarnoja, Zhou, Abbeel, and Levine]{Haarnoja18a}
T.~Haarnoja, A.~Zhou, P.~Abbeel, and S.~Levine.
\newblock Soft actor-critic: Off-policy maximum entropy deep reinforcement
  learning with a stochastic actor.
\newblock In \emph{International Conference on Machine Learning}, pp.\
  1856--1865, 2018.

\bibitem[Hausknecht \& Stone(2015)Hausknecht and Stone]{Hausknecht15a}
M.~Hausknecht and P.~Stone.
\newblock Deep recurrent {Q}-learning for partially observable {MDPs}.
\newblock In \emph{AAAI Conference on Artificial Intelligence}, 2015.

\bibitem[{Hernandez-Gardiol} \& Mahadevan(2000){Hernandez-Gardiol} and
  Mahadevan]{HernandezGardio00a}
N.~{Hernandez-Gardiol} and S.~Mahadevan.
\newblock Hierarchical memory-based reinforcement learning.
\newblock In \emph{Advances in Neural Information Processing Systems}, 2000.

\bibitem[Jiang et~al.(2018)Jiang, Ekwedike, and Liu]{Jiang18a}
D.~R. Jiang, E.~Ekwedike, and H.~Liu.
\newblock Feedback-based tree search for reinforcement learning.
\newblock In \emph{International Joint Conference on Artificial Intelligence},
  pp.\  2284--2293, 2018.

\bibitem[Kaelbling et~al.(1996)Kaelbling, Littman, and Moore]{kaelbling96a}
L.~P. Kaelbling, M.~L. Littman, and A.~W. Moore.
\newblock Reinforcement learning: A survey.
\newblock \emph{Journal of AI Research}, 4:\penalty0 237–--285, 1996.

\bibitem[Kaelbling et~al.(1998)Kaelbling, Littman, and Cassandra]{Kaelbling98a}
L.~P. Kaelbling, M.~L. Littman, and A.~R. Cassandra.
\newblock Planning and acting in partially observable stochastic domains.
\newblock \emph{Artificial Intelligence}, 101\penalty0 (1-2):\penalty0 99--134,
  1998.

\bibitem[Kakade(2002)]{kakade02a}
S.~M. Kakade.
\newblock A natural policy gradient.
\newblock In \emph{Advances in Neural Information Processing Systems},
  volume~14. MIT Press, 2002.

\bibitem[Kamigaito et~al.(2021)Kamigaito, Zhang, Takamura, and
  Okumura]{Kamigaito21a}
H.~Kamigaito, P.~Zhang, H.~Takamura, and M.~Okumura.
\newblock An empirical study of generating texts for search engine advertising.
\newblock In \emph{Conference of the North American Chapter of the Association
  for Computational Linguistics: Industry Papers}, pp.\  255--262, 2021.

\bibitem[Kiegeland \& Kreutzer(2021)Kiegeland and Kreutzer]{Kiegeland21a}
S.~Kiegeland and J.~Kreutzer.
\newblock Revisiting the weaknesses of reinforcement learning for neural
  machine translation.
\newblock In \emph{Annual Conference of the North American Chapter of the
  Association for Computational Linguistics}, 2021.

\bibitem[Kim et~al.(2020)Kim, Lee, Lim, Kaelbling, and
  Lozano-P\'{e}rez]{Kim20a}
B.~Kim, K.~Lee, S.~Lim, L.~P. Kaelbling, and T.~Lozano-P\'{e}rez.
\newblock {Monte Carlo} tree search in continuous spaces using {Voronoi}
  optimistic optimization with regret bounds.
\newblock In \emph{AAAI Conference on Artificial Intelligence}, 2020.

\bibitem[Kimura et~al.(1997)Kimura, Miyazaki, and Kobayashi]{kimura97a}
H.~Kimura, K.~Miyazaki, and S.~Kobayashi.
\newblock Reinforcement learning in {POMDPs} with function approximation.
\newblock In \emph{International Conference on Machine Learning}, 1997.

\bibitem[Kocsis \& Szepesv\'ari(2006)Kocsis and Szepesv\'ari]{Kocsis06a}
L.~Kocsis and C.~Szepesv\'ari.
\newblock Bandit based {M}onte-{C}arlo planning.
\newblock In \emph{European Conference on Machine Learning}, pp.\  282--293,
  2006.

\bibitem[Kuncheva(2014)]{Kuncheva14a}
L.~I. Kuncheva.
\newblock \emph{Combining Pattern Classifiers: Methods and Algorithms, 2nd
  Edition}.
\newblock John Wiley and Sons, 2014.

\bibitem[Lattimore \& Szepesv\'ari(2020)Lattimore and
  Szepesv\'ari]{Lattimore20a}
T.~Lattimore and C.~Szepesv\'ari.
\newblock \emph{Bandit Algorithms}.
\newblock Cambridge University Press, 2020.

\bibitem[Li et~al.(2011)Li, Littman, Walsh, and Strehl]{Li11a}
L.~Li, M.~L. Littman, T.~J. Walsh, and A.~L. Strehl.
\newblock Knows what it knows: A framework for self-aware learning.
\newblock \emph{Machine Learning}, 82\penalty0 (3):\penalty0 399--443, 2011.

\bibitem[Liu et~al.(2022)Liu, Lei, Lv, and Zhou]{Liu22a}
X.~Liu, W.~Lei, J.~Lv, and J.~Zhou.
\newblock Abstract rule learning for paraphrase generation.
\newblock In \emph{International Joint Conference on Artificial Intelligence},
  2022.

\bibitem[Loch \& Singh(1998)Loch and Singh]{Loch98a}
J.~Loch and S.~Singh.
\newblock Using eligibility traces to find the best memoryless policy in
  partially observable {M}arkov decision processes.
\newblock In \emph{International Conference on Machine Learning}, 1998.

\bibitem[Majeed \& Hutter(2018)Majeed and Hutter]{Majeed18a}
S.~J. Majeed and M.~Hutter.
\newblock On {Q}-learning convergence for non-{M}arkov decision processes.
\newblock In \emph{International Joint Conference on Artificial Intelligence},
  pp.\  2546--2552, 2018.

\bibitem[Mansley et~al.(2011)Mansley, Weinstein, and Littman]{Mansley11a}
C.~Mansley, A.~Weinstein, and M.~Littman.
\newblock Sample-based planning for continuous action markov decision
  processes.
\newblock In \emph{International Conference on Automated Planning and
  Scheduling}, 2011.

\bibitem[Mao et~al.(2020)Mao, Zhang, Xie, and Ba\c{s}ar]{Mao20a}
W.~Mao, K.~Zhang, Q.~Xie, and T.~Ba\c{s}ar.
\newblock {POLY-HOOT}: {Monte-Carlo} planning in continuous space mdps with
  non-asymptotic analysis.
\newblock In \emph{Advances in Neural Information Processing Systems}, 2020.

\bibitem[Morimura et~al.(2014)Morimura, Osogami, and Shirai]{morimura14:aaai}
T.~Morimura, T.~Osogami, and T.~Shirai.
\newblock Mixing-time regularized policy gradient.
\newblock In \emph{AAAI Conference on Artificial Intelligence}, 2014.

\bibitem[Ouyang et~al.(2022)Ouyang, Wu, Jiang, Almeida, Wainwright, Mishkin,
  Zhang, Agarwal, Slama, Ray, Schulman, Hilton, Kelton, Miller, Simens, Askell,
  Welinder, Christiano, Leike, and Lowe]{instructGPT}
L.~Ouyang, J.~Wu, X.~Jiang, D.~Almeida, C.~Wainwright, P.~Mishkin, C.~Zhang,
  S.~Agarwal, K.~Slama, A.~Ray, J.~Schulman, J.~Hilton, F.~Kelton, L.~Miller,
  M.~Simens, A.~Askell, P.~Welinder, P.~Christiano, J.~Leike, and R.~Lowe.
\newblock Training language models to follow instructions with human feedback.
\newblock In \emph{Advances in Neural Information Processing Systems},
  volume~35, pp.\  27730--27744, 2022.

\bibitem[Paulus et~al.(2018)Paulus, Xiong, and Socher]{Paulus18a}
R.~Paulus, C.~Xiong, and R.~Socher.
\newblock A deep reinforced model for abstractive summarization.
\newblock In \emph{International Conference on Learning Representations}, 2018.

\bibitem[Peters \& Schaal(2008)Peters and Schaal]{Peters08b}
J.~Peters and S.~Schaal.
\newblock Reinforcement learning of motor skills with policy gradients.
\newblock \emph{Neural Networks}, 21\penalty0 (4):\penalty0 682--697, 2008.

\bibitem[Poupart \& Vlassis(2008)Poupart and Vlassis]{Poupart08a}
P.~Poupart and N.~Vlassis.
\newblock Model-based {Bayesian} reinforcement learning in partially observable
  domains.
\newblock In \emph{International Symposium on Artificial Intelligence and
  Mathematics}, 2008.

\bibitem[Puterman(1994)]{Puterman94a}
M.~L. Puterman.
\newblock \emph{{M}arkov Decision Processes: Discrete Stochastic Dynamic
  Programming}.
\newblock John Wiley and Sons, 1994.

\bibitem[Qin et~al.(2023)Qin, Gao, Li, Zhu, and Xie]{Qin23a}
A.~Qin, F.~Gao, Q.~Li, S.-C. Zhu, and S.~Xie.
\newblock Learning non-markovian decision-making from state-only sequences.
\newblock In \emph{Advances in Neural Information Processing Systems}, 2023.

\bibitem[Rennie et~al.(2017)Rennie, Marcheret, Mroueh, Ross, and Goel]{SCST}
S.~J. Rennie, E.~Marcheret, Y.~Mroueh, J.~Ross, and V.~Goel.
\newblock Self-critical sequence training for image captioning.
\newblock In \emph{IEEE Conference on Computer Vision and Pattern Recognition
  (CVPR)}, pp.\  1179--1195, 2017.

\bibitem[Schrittwieser et~al.(2020)Schrittwieser, Antonoglou, Hubert, Simonyan,
  Sifre, Schmitt, Guez, Lockhart, Hassabis, Graepel, Lillicrap, and
  Silver]{Schrittwieser20a}
J.~Schrittwieser, I.~Antonoglou, T.~Hubert, K.~Simonyan, L.~Sifre, S.~Schmitt,
  A.~Guez, E.~Lockhart, D.~Hassabis, T.~Graepel, T.~Lillicrap, and D.~Silver.
\newblock Mastering {A}tari, go, chess and shogi by planning with a learned
  model.
\newblock In \emph{Nature}, volume 588, 2020.

\bibitem[Schulman et~al.(2015)Schulman, Levine, Abbeel, Jordan, and
  Moritz]{Schulman15a}
J.~Schulman, S.~Levine, P.~Abbeel, M.~Jordan, and P.~Moritz.
\newblock Trust region policy optimization.
\newblock In \emph{International Conference on Machine Learning}, pp.\
  1889--1897, 2015.

\bibitem[Schulman et~al.(2017)Schulman, Wolski, Dhariwal, Radford, and
  Klimov]{Schulman17a}
J.~Schulman, F.~Wolski, P.~Dhariwal, A.~Radford, and O.~Klimov.
\newblock Proximal policy optimization algorithms.
\newblock In \emph{arXiv preprint arXiv:1707.06347}, 2017.

\bibitem[Silver \& Veness(2010)Silver and Veness]{Silver10a}
D.~Silver and J.~Veness.
\newblock {Monte-Carlo} planning in large {POMDPs}.
\newblock In \emph{Advances in Neural Information Processing Systems},
  volume~23, pp.\  2164–--2172, 2010.

\bibitem[Silver et~al.(2016)Silver, Huang, Maddison, Guez, Sifre, van~den
  Driessche, Schrittwieser, Antonoglou, Panneershelvam, Lanctot, Dieleman,
  Grewe, Nham, Kalchbrenner, Sutskever, Lillicrap, Leach, Kavukcuoglu, Graepel,
  and Hassabis]{Silver16a}
D.~Silver, A.~Huang, C.~J. Maddison, A.~Guez, L.~Sifre, G.~van~den Driessche,
  J.~Schrittwieser, I.~Antonoglou, V.~Panneershelvam, M.~Lanctot, S.~Dieleman,
  D.~Grewe, J.~Nham, N.~Kalchbrenner, I.~Sutskever, T.~Lillicrap, M.~Leach,
  K.~Kavukcuoglu, T.~Graepel, and D.~Hassabis.
\newblock Mastering the game of {Go} with deep neural networks and tree search.
\newblock \emph{Nature}, 529\penalty0 (7587):\penalty0 484--489, 2016.

\bibitem[Silver et~al.(2017{\natexlab{a}})Silver, Hubert, Schrittwieser,
  Antonoglou, Lai, Guez, Lanctot, Sifre, Kumaran, Graepel, Lillicrap, Simonyan,
  and Hassabis]{Silver17b}
D.~Silver, T~Hubert, J.~Schrittwieser, I~Antonoglou, M.~Lai, A~Guez,
  M.~Lanctot, L~Sifre, D.~Kumaran, T.~Graepel, T.~Lillicrap, K.~Simonyan, and
  D.~Hassabis.
\newblock Mastering chess and shogi by self-play with a general reinforcement
  learning algorithm.
\newblock \emph{arXiv preprint arXiv:1712.01815}, 2017{\natexlab{a}}.

\bibitem[Silver et~al.(2017{\natexlab{b}})Silver, Schrittwieser, Simonyan,
  Antonoglou, Huang, Guez, Hubert, Baker, Lai, Bolton, Chen, Lillicrap, Hui,
  Sifre, Driessche, Graepel, and Hassabis]{Silver17a}
D.~Silver, J.~Schrittwieser, K.~Simonyan, I.~Antonoglou, A.~Huang, A.~Guez,
  T.~Hubert, L.~Baker, M.~Lai, A.~Bolton, Y.~Chen, T.~Lillicrap, F.~Hui,
  L.~Sifre, G.~{van den} Driessche, T.~Graepel, and D.~Hassabis.
\newblock Mastering the game of {Go} without human knowledge.
\newblock \emph{Nature}, 550\penalty0 (7676):\penalty0 354–--359,
  2017{\natexlab{b}}.

\bibitem[Singh et~al.(2012)Singh, James, and Rudary]{Singh12a}
S.~Singh, M.~James, and M.~Rudary.
\newblock Predictive state representations: A new theory for modeling dynamical
  systems.
\newblock In \emph{Conference on Uncertainty in Artificial Intelligence}, 2012.

\bibitem[Soemers et~al.(2019)Soemers, Piette, Stephenson, and
  Browne]{Soemers19a}
D.~J. N.~J. Soemers, \'{E}. Piette, M.~Stephenson, and C.~Browne.
\newblock Learning policies from self-play with policy gradients and {MCTS}
  value estimates.
\newblock In \emph{IEEE Conference on Games}, 2019.

\bibitem[Sondik(1971)]{sondikthesis}
E.~J. Sondik.
\newblock \emph{The optimal control of partially observable {M}arkov
  processes}.
\newblock PhD thesis, Stanford University, 1971.

\bibitem[Sutton \& Barto(2018)Sutton and Barto]{sutton2nd}
R.~S. Sutton and A.~G. Barto.
\newblock \emph{Reinforcement Learning}.
\newblock MIT Press, 2nd edition, 2018.

\bibitem[\'{S}wiechowski et~al.(2021)\'{S}wiechowski, Godlewski, Sawicki, and
  Ma\'{n}dziuk]{Swiechowski21a}
M.~\'{S}wiechowski, K.~Godlewski, B.~Sawicki, and J.~Ma\'{n}dziuk.
\newblock Monte carlo tree search: A review of recent modifications and
  applications.
\newblock In \emph{arXiv preprint arXiv:2103.04931}, 2021.

\bibitem[Tang et~al.(2017)Tang, Houthooft, Foote, Stooke, Chen, Y.Duan,
  Schulman, DeTurck, and Abbeel]{Tang17a}
H.~Tang, R.~Houthooft, D.~Foote, A.~Stooke, X.~Chen, Y.Duan, J.~Schulman,
  F.~DeTurck, and P.~Abbeel.
\newblock {\#Exploration}: A study of count-based exploration for deep
  reinforcement learning.
\newblock In \emph{Advances in Neural Information Processing Systems}, pp.\
  2750--2759, 2017.

\bibitem[Thi{\'{e}}baux et~al.(2006)Thi{\'{e}}baux, Gretton, Slaney, Price, and
  Kabanza]{Thiebaux06a}
S.~Thi{\'{e}}baux, C.~Gretton, J.~Slaney, D.~Price, and F.~Kabanza.
\newblock Decision-theoretic planning with non-markovian rewards.
\newblock \emph{Journal of Artiﬁcial Intelligence Research}, 25:\penalty0
  17--74, 2006.

\bibitem[Vodopivec et~al.(2017)Vodopivec, Samothrakis, and
  \u{S}ter]{Vodopivec17a}
T.~Vodopivec, S.~Samothrakis, and B.~\u{S}ter.
\newblock On monte carlo tree search and reinforcement learning.
\newblock \emph{Journal of Artificial Intelligence Research}, 60:\penalty0
  881--936, 2017.

\bibitem[Wang et~al.(2021)Wang, Du, Zhu, Ke, Chen, Hao, and Wang]{Wang21a}
X.~Wang, Y.~Du, S.~Zhu, L.~Ke, Z.~Chen, J.~Hao, and J.~Wang.
\newblock Ordering-based causal discovery with reinforcement learning.
\newblock In \emph{International Joint Conference on Artificial Intelligence},
  2021.

\bibitem[Whitehead \& Lin(1995)Whitehead and Lin]{Whitehead95a}
S.~D. Whitehead and L.~J. Lin.
\newblock Reinforcement learning of non-{M}arkov decision processes.
\newblock \emph{Artificial Intelligence}, 73\penalty0 (1-2):\penalty0 271--306,
  1995.

\bibitem[Wierstra et~al.(2010)Wierstra, Förster, Peters, and
  Schmidhuber]{Wierstra10a}
D.~Wierstra, A.~Förster, J.~Peters, and J.~Schmidhuber.
\newblock Recurrent policy gradients.
\newblock \emph{Logic Journal of the IGPL}, 18\penalty0 (5):\penalty0
  620–--634, 2010.

\bibitem[Williams(1992)]{williams92a}
R.~J. Williams.
\newblock Simple statistical gradient-following algorithms for connectionist
  reinforcement learning.
\newblock \emph{Machine Learning}, 8:\penalty0 229--256, 1992.

\bibitem[Xia et~al.(2020)Xia, Zhou, Shi, Lu, and Huang]{Xia20a}
Y.~Xia, J.~Zhou, Z.~Shi, C.~Lu, and H.~Huang.
\newblock Generative adversarial regularized mutual information policy gradient
  framework for automatic diagnosis.
\newblock In \emph{AAAI Conference on Artificial Intelligence}, 2020.

\bibitem[Xiao et~al.(2019)Xiao, Huang, Mei, Schuurmans, and
  M{\"{u}}ller]{Xiao19a}
C.~Xiao, R.~Huang, J.~Mei, D.~Schuurmans, and M.~M{\"{u}}ller.
\newblock Maximum entropy {M}onte-{C}arlo planning.
\newblock In \emph{Advances in Neural Information Processing Systems}, 2019.

\bibitem[You et~al.(2018)You, Liu, Ying, Pande, and Leskovec]{You18a}
J.~You, B.~Liu, R.~Ying, V.~Pande, and J.~Leskovec.
\newblock Graph convolutional policy network for goal-directed molecular graph
  generation.
\newblock In \emph{Advances in Neural Information Processing Systems}, 2018.

\bibitem[Young et~al.(2013)Young, Ga\u{s}i\'c, Thomson, and Williams]{Young13a}
S.~Young, M.~Ga\u{s}i\'c, B.~Thomson, and J.~D. Williams.
\newblock {POMDP}-based statistical spoken dialog systems: A review.
\newblock In \emph{Proceedings of the IEEE}, volume 101, pp.\  1160--1179.
  IEEE, 2013.

\bibitem[Yu et~al.(2017)Yu, Zhang, Wang, and Yu]{SeqGAN}
L.~Yu, W.~Zhang, J.~Wang, and Y.~Yu.
\newblock Seqgan: Sequence generative adversarial nets with policy gradient.
\newblock In \emph{AAAI Conference on Artificial Intelligence}, 2017.

\bibitem[Yu et~al.(2011)Yu, Zhou, Chan, Chen, and Yang]{Yu11a}
T.~Yu, B.~Zhou, K.~W. Chan, L.~Chen, and B.~Yang.
\newblock Stochastic optimal relaxed automatic generation control in
  non-{M}arkov environment based on multi-step {Q}($lambda$) learning.
\newblock \emph{IEEE Transactions on Power Systems}, 26\penalty0 (3):\penalty0
  1272 -- 1282, 2011.

\bibitem[Zhang et~al.(2021)Zhang, Kim, {O'D}onoghue, and Boyd]{Zhang21a}
J.~Zhang, J.~Kim, B.~{O'D}onoghue, and S.~Boyd.
\newblock Sample efficient reinforcement learning with reinforce.
\newblock In \emph{AAAI Conference on Artificial Intelligence}, 2021.

\bibitem[Zheng et~al.(2018)Zheng, Oh, and S]{Zheng18a}
Z.~Zheng, J.~Oh, and Singh S.
\newblock On learning intrinsic rewards for policy gradient methods.
\newblock In \emph{Advances in Neural Information Processing Systems},
  volume~31, 2018.

\bibitem[Zhou et~al.(2018)Zhou, Xiong, Socher, and Socher]{Zhou18a}
Y.~Zhou, C.~Xiong, R.~Socher, and R.~Socher.
\newblock Improving end-to-end speech recognition with policy learning.
\newblock In \emph{IEEE International Conference on Acoustics, Speech and
  Signal Processing}, 2018.

\end{thebibliography}
